\documentclass[11pt]{article}

\ifdefined\pdfobjcompresslevel
\fi

\usepackage{lmodern}
\usepackage{pdfpages}

\usepackage[preprint]{acl}

\usepackage{times}
\usepackage{latexsym}
\usepackage{supertabular} 
\usepackage{booktabs}
\usepackage{amsfonts,amsmath,amssymb,amsthm}
\usepackage{multirow}
\usepackage{booktabs}
\usepackage[table]{xcolor}
\usepackage{array} 
\usepackage{inconsolata}
\usepackage{graphicx}
\usepackage{adjustbox}
\usepackage{amsfonts}
\usepackage{amssymb}
\usepackage{colortbl}
\usepackage{svg}
\usepackage[T1]{fontenc}
\usepackage{url}

\usepackage[utf8]{inputenc}

\usepackage{microtype}

\usepackage{inconsolata}

\usepackage{graphicx}
\usepackage{fontawesome5}

\title{The Necessity of Setting Temperature in LLM-as-a-Judge}

\author{
 \textbf{Lujun Li\textsuperscript{$\spadesuit$1}},
 \textbf{Lama Sleem\textsuperscript{$\spadesuit$1}},
 \textbf{Yangjie Xu\textsuperscript{$\spadesuit$1}},
 \\
 \textbf{Yewei Song\textsuperscript{1}},
  \textbf{Aolin Jia\textsuperscript{2}},
 \textbf{Jerome Francois\textsuperscript{1}},
 \textbf{Radu State\textsuperscript{1}},
\\
 \textsuperscript{1}University of Luxembourg,
 \textsuperscript{2}ETH Zürich
\\
}

\usepackage{times}
\usepackage{latexsym}
\usepackage[T1]{fontenc}
\usepackage{inconsolata}
\usepackage{tcolorbox}
\tcbuselibrary{skins,breakable,xparse}

\newtcolorbox{promptbox}[2][]{
  enhanced,
  breakable,
  colback=gray!10,
  colframe=gray!70,
  coltitle=white,
  fonttitle=\bfseries\small\ttfamily\color{white},
  fontupper=\ttfamily\footnotesize,
  boxrule=0.8pt,
  title={#2},
  title style={fill=gray!70},
  left=5mm, right=5mm,
  top=3mm, bottom=3mm,
  arc=2mm,
  borderline west={3pt}{0pt}{gray!80},
  boxsep=2pt,
  #1
}

\newtcolorbox{userbox}[2][]{
  enhanced,
  breakable,
  colback=blue!3,
  colframe=blue!60!black,
  coltitle=white,
  fonttitle=\bfseries\small\ttfamily\color{white},
  fontupper=\ttfamily\footnotesize,
  boxrule=0.8pt,
  title={#2},
  title style={fill=blue!60!black},
  left=5mm, right=5mm,
  top=3mm, bottom=3mm,
  arc=2mm,
  borderline west={3pt}{0pt}{blue!70},
  boxsep=2pt,
  #1
}

\newtcolorbox{responsebox}[2][]{
  enhanced,
  breakable,
  colback=green!5,
  colframe=green!65!black,
  coltitle=white,
  fonttitle=\bfseries\small\ttfamily\color{white},
  fontupper=\ttfamily\footnotesize,
  boxrule=0.8pt,
  title={#2},
  title style={fill=green!65!black},
  left=5mm, right=5mm,
  top=3mm, bottom=3mm,
  arc=2mm,
  borderline west={3pt}{0pt}{green!75},
  boxsep=2pt,
  #1
}

\definecolor{lightblue}{RGB}{230,240,255}
\definecolor{lightred}{RGB}{255,230,230}

\definecolor{TColdBg}{RGB}{233,245,255} 
\definecolor{TMidBg}{RGB}{236,247,239}  
\definecolor{THotBg}{RGB}{255,242,232}  
\definecolor{TColdText}{RGB}{0,92,171}
\definecolor{TMidText}{RGB}{20,115,60}
\definecolor{THotText}{RGB}{176,60,0}

\begin{document}
 \maketitle
\begin{abstract}
Large Language Models (LLMs) are increasingly used as automated judges, yet decoding temperature is often set by intuition rather than evidence. We run over one million large scale evaluations, multiple open-source models, prompting strategies, and judging paradigms to examine how temperature affects judge reliability. Low temperatures yield consistent outputs with few errors, especially in smaller models, while higher temperatures reduce consistency and increase errors. Yet consistency alone does not guarantee sound judgments: low temperatures can narrow the range of reasoning explored. We characterize judge behavior through four reasoning personas and find that temperature shifts their distributions in model-dependent ways, most visibly in ambiguous cases where competing answers are hard to distinguish. Repeated high-temperature sampling reveals this hidden uncertainty, and combining it with majority voting or using high temperature for reasoning and low temperature for the final decision, improves agreement on ambiguous cases without hurting clear ones, at the cost of extra inference. Our findings suggest a simple rule: keep temperature low for stable outputs, but explore more when the judgment is hard.

\end{abstract}

\section{Introduction}

With the widespread adoption of large language models (LLMs) in domains such as text generation, question-answering systems, and automated evaluation, assessing the output quality of LLMs has emerged as a pivotal challenge \cite{gu2025surveyllmasajudge}. Recent trends indicate that employing LLMs as evaluators or judges---commonly termed \textbf{LLM-as-a-Judge}---is progressively supplanting common human-annotated evaluation. This shift is driven by the capacity of model-based evaluation to substantially reduce human labeling costs, accelerate assessment cycles, and demonstrate superior consistency in sophisticated tasks like multi-turn dialogues and open-ended question answering \cite{tang2025learning}. Nevertheless, existing research reveals that LLM-based evaluation outcomes are susceptible to influences from generation parameters, such as \textbf{sampling temperature}~\cite{du2025optimizingtemperaturelanguagemodels}, which can induce score volatility, positional bias and decreased consistency, thus undermining the interpretability and reliability of the assessments.

\begin{figure}[htbp]
\hspace{-0.5cm}
    \includegraphics[width=1.15\linewidth]{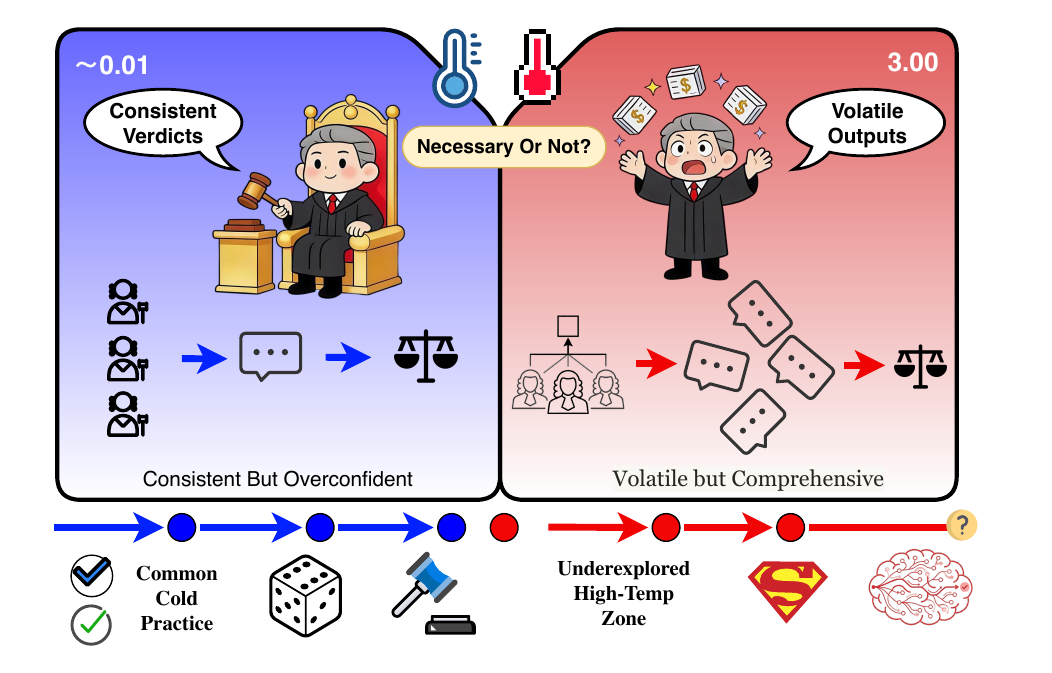}
  \caption{LLM-as-a-Judge Paradigm}
  \label{fig:llmasaaj}
\end{figure}

To the best of our knowledge, empirical settings for the LLM temperature---such as values close to 0---are commonly employed in LLM-as-a-Judge paradigms. This practice is well-understood, particularly given cost considerations and reproducibility, as more powerful models have been demonstrated to exhibit exceptionally high alignment with human judgments \cite{10.5555/3666122.3668142}. However, a critical question remains: \textbf{Does decoding temperature influence the evaluative performance of LLM judges, and how should it be configured to balance reliability with reasoning diversity?} Moreover, the extent to which different LLM-as-a-judge configurations are affected represents a still underexplored avenue in real-world applications scenarios. Therefore, although the role of temperature control in the diversity of generation \cite{peeperkorn2024temperaturecreativityparameterlarge} has been extensively investigated, its underlying mechanisms on the \textbf{``judging behavior''} remain underexplored through a systematic analysis. 

This work makes three main connected contributions:
\begin{enumerate}
    \item \textbf{Large-scale empirical study of temperature.} We conduct a large-scale empirical study of the temperature parameter, leveraging two datasets and multiple state-of-the-art open-source models. We execute over one million evaluations across a wide range of temperature configurations, spanning diverse prompting strategies, LLM architectures, and judging paradigms. The results provide solid statistical evidence and practical insights, showing that decoding temperature is a critical tuning knob---one that meaningfully shapes the stability, format compliance, and reproducibility of evaluation outcomes.
    \item \textbf{Behavioral analysis of LLM judges.} Through case studies of LLM judges' outputs, we uncover deep insights into their behavioral dynamics under different settings, revealing that temperature genuinely shifts the judge's \emph{persona}.
    \item \textbf{Thermo-Judge for ambiguous cases.} We further probe how higher temperatures can boost judging performance by evaluating ``Thermo-Judge'' on a set of ambiguous cases, and explore token-level temperature annealing method for stable gains---validating the value of diverse temperature settings in LLM-as-a-Judge.
\end{enumerate}

\section{Related work}

\noindent \textbf{Commonly used but understudied:} Although LLMs have different types of bias \cite{ye2024justiceprejudicequantifyingbiases,shi-etal-2025-judging,chen-etal-2024-humans}, LLMs-as-a-Judge have emerged as a novel form of annotator capable of performing classification tasks across diverse text-related domains~\cite{GU2026101253}. \textbf{Temperature settings have already been widely adopted} in LLM-as-a-Judge frameworks, with substantial variation across cases, typically ranging from 0.01 to 1.0. Fundamentally, they remain deep learning classifiers, with their in-context learning generalization enabling evaluations tailored to varied requirements. \citeauthor{10.5555/3666122.3668142} validated the high alignment between LLMs-as-a-Judge and human judgments through extensive manual inspections and experiments across multiple dimensions, employing a fixed temperature setting of 0.7. \citeauthor{shi-etal-2025-judging} conducted an in-depth investigation into position bias under diverse configurations by introducing metrics such as repeated stability and position consistency, underscoring the necessity of swapping candidate pair positions; however, this study only adopted a single temperature of 1.0. Another work~\cite{wei2025systematicevaluationllmasajudgellm} preliminarily examined the impact of temperature on LLMs-as-a-Judge, testing values from 0.0 to 0.7 and finding peak self-consistency and accuracy at 0.1; nevertheless, the overall temperature effects were not pronounced, the range was limited, and the exploration remained insufficiently comprehensive.

\noindent \textbf{T affects Judging:}  However, \citeauthor{LI2025242} provides preliminary evidence that higher temperature settings can affect LLM outputs and may even improve performance in some cases. Related studies further suggest that temperature can substantially influence the behavior of LLMs \cite{renze-2024-effect, deepseek_api_temp_2026}, raising the possibility that it may also affect the effectiveness of LLM-as-a-Judge systems. However, prior explorations of temperature effects remain limited in both breadth and depth and whether distinct temperature settings are required for LLM-as-a-Judge systems to optimize judgment outcomes remains an open question.

\section{Methods}

\subsection{Definition}

Let \( v \in \mathcal{V} \) denote the final verdict in the discrete verdict space \( \mathcal{V} \) \cite{ACKLEY1985147}.
Given a question \( q \), we consider an input tuple \( (x, r) \), where \( x \) denotes the model-generated answer to \( q \), and \( r \) denotes the reference (gold) answer when available. In settings without reference answers, the judgment is performed solely based on \( x \).

\begin{equation}
\begin{aligned}
p_T(v = i \mid x, r)
&= \frac{\exp\!\left(a_i(x, r)/T\right)}
{\sum_{j \in \mathcal{V}} \exp\!\left(a_j(x, r)/T\right)}, \\
\hat{v}
&= \mathrm{Decode}\!\left(p_T(\cdot \mid x, r)\right).
\end{aligned}
\end{equation}

Here, \( a_i(x, r) \) denotes the logit associated with verdict \( i \), and \( T > 0 \) is the decoding temperature. When the reference answer \( r \) is not available, the formulation naturally reduces to conditioning on \( x \) alone. The temperature rescales the logits to control the sharpness of the verdict distribution without altering their relative order, and therefore does not change the ranking of candidate output tokens. In practice, this implies that temperature primarily affects decision variance and reproducibility, rather than inducing substantial shifts in average accuracy. Furthermore, due to the autoregressive nature of decoding, LLMs generation typically adhere to a Markovian inference process\cite{zhang2026markovcategoricalframeworklanguage}, with output diversity governed by a random seed 
that controls the initial generation; however, incorporating an initial reasoning 
phase—such as Chain-of-Thought (CoT) prompting \citep{DBLP:journals/corr/abs-2201-11903}—can lead to fundamentally divergent 
model outputs.

\subsection{Experimental Design}

In this study, we examine six temperature settings: \([0.01, 0.5, 1.0, 1.5, 2.0, 3.0]\). 
To reduce stochastic variability, each experimental condition is repeated ten times using different random seeds. We employ \texttt{Qwen3-Next-80B-A3B-Instruct-FP8 (Qwen3-Next-80B)}~\cite{qwen2.5-1m, qwen3technicalreport} and \texttt{gemma-3-27b-it (gemma-3-27b)}~\cite{gemma_2025} as the primary open-weight judge models, supplemented with additional models for robustness checks and auxiliary analyses, to assess whether temperature-induced effects generalize across models. The temperature range follows prior work~\cite{LI2025242}, which also reports that \texttt{top\_p} and \texttt{top\_k} have limited effects on LLMs' capabilities. Accordingly, we fix \texttt{top\_p} at 0.95 and the maximum generation length at 2048 tokens in all experiments.

\paragraph{Judge Types} Following the taxonomy of judge types proposed by \cite{10.5555/3666122.3668142}, we examine three representative LLM-as-a-Judge paradigms: \textbf{pairwise comparison}, \textbf{single-answer grading}, and \textbf{reference-guided grading}. Pairwise comparison requires the model to select the better response between two candidates; single-answer grading assigns an absolute score or label to a single response; and reference-guided grading evaluates a response with respect to a provided gold answer (see prompt details in Appendix~\ref{appendix:prompt template}). Each paradigm is tested using both the direct base prompt and the CoT prompt, as these represent the most common approaches in LLM-as-a-judge evaluations. 
We do not additionally test few-shot prompting here, primarily due to its potential to introduce unclear biases whose interpretability remains underexplored. 
Furthermore, while positional bias (i.e., LLM preferences when options are presented first versus last) is a compelling topic, it is excluded from this study, as our primary focus is the impact of temperature on the judge performance. In addition, we conduct supplementary experiments on a new ensemble judge method to demonstrate the complementary role of higher temperatures in certain scenarios.

\paragraph{Personas Definitions}

We define temperature-induced judge personas as recurring behavioral profiles in judge outputs---not fixed personality types. Building on prior work on sampling temperature, judge biases, and multidimensional evaluation, we identify four mutually exclusive personas along reasoning breadth, evidence coverage, and decision certainty. The \textbf{Expedient Judge\cite{renze-2024-effect}} gives a quick, direct verdict with narrow evidence coverage. The \textbf{Salient Judge\cite{ye2024justiceprejudicequantifyingbiases,li2024llmsasjudgescomprehensivesurveyllmbased}} fixates on a single salient flaw, vulnerability, risk, or bias while largely ignoring other dimensions. The \textbf{Comprehensive Judge\cite{peeperkorn2024temperaturecreativityparameterlarge}} weighs at least two criteria, evidence sources, or perspectives and lands on a clear, stable verdict. The \textbf{Uncertain Judge\cite{hobelsberger2025systematicevaluationuncertaintyestimation}} explores broadly but hedges, produces a verdict that contradicts its own reasoning, or flips across repeated samples. To label personas at scale, a strong reference model (Qwen3.8-Flash-Next) serves as an auxiliary annotator under a fixed rubric and tie-breaking rules. It sees the full chain-of-thought but not the temperature or other experimental settings, so labels capture \emph{how} the judge decided---where temperature effects are most visible.

\subsection{Dataset}

For our dataset selection, we primarily employ \textbf{MT-Bench Human Judgments} \cite{10.5555/3666122.3668142}, which encompasses extensive data from models such as GPT-3.5-Turbo, Vicuna-13B-v1.2, Llama-13B, and others, all derived from the MT-Bench dataset~\cite{10.5555/3666122.3668142}. Its main advantage lies in the inclusion of \textbf{human-verified annotations}, which allow us to assess the alignment between LLM-as-a-Judge performance and human evaluators under different temperature settings. From this dataset, we randomly sampled 350 questions to construct our temperature test set.

Since MT-Bench was released in late 2023, we supplement it with \textbf{MMLU-Pro}~\cite{10.5555/3737916.3740934}, which offers substantially greater difficulty than the original MMLU and recommends chain-of-thought reasoning for all tasks, making it well suited for assessing LLM-as-a-Judge performance on complex reasoning outcomes. We sample 150 items from the original test set following the dataset's category proportions and randomly select three open-source models---Llama-3.1-8B-Instruct, Gemma-3-12B-it, and Qwen2.5-14B-Instruct---to generate responses for judging. As MMLU-Pro does not include human preference annotations, we additionally collect reference preference labels using GPT-5.4 (state-of-the-art as of February 2026) under a fixed judging protocol (see prompt details in Appendix~\ref{appendix:judge_prompt_4.5}). Our open-source dataset thus comprises 500 questions, responses, and judge selections, combining human-verified annotations where available with reference-based labels elsewhere, enabling a systematic investigation of temperature's impact on LLM-as-a-Judge performance.

\subsection{Metrics}

Three metrics are used to evaluate the judge behavior: agreement rate, consistency, and error rate. 
\textbf{Agreement Rate (AGR)} measures the extent to which the decisions produced by the LLM judge align with human-verified reference labels (i.e., consistency with the decision made by the ``judge in the court''), and thus serves as a primary indicator of judge performance.
$$
\text{AGR} = \frac{1}{N}\sum_{i=1}^N \mathbf{1}(\hat{y}_i = y_i^*) \cdot \mathbf{1}(\hat{y}_i \neq \emptyset \wedge y_i^* \neq \emptyset)
$$
where $\hat{y}_i\in\{\text{model\_a, model\_b, tie}\}$ is the normalized prediction, $y_i^*$ is human gold judge standard from the tested datasets and $\mathbf{1}$ is an indicator function. \textbf{Consistency (CON)} uses 1-Flip Consistency, which measures judge stability across repeated runs under identical experimental conditions by computing the proportion of adjacent repeats with consistent judgments:
$$
\text{CON}_q = 1 - \frac{1}{n_q-1}\sum_{k=1}^{n_q-1} \mathbf{1}(\hat{y}_{q,k} \neq \hat{y}_{q,k+1})
$$
where \( n_q \) denotes the number of repeated evaluations for question \( q \), and \( \hat{y}_{q,k} \) is the predicted judgment at the \( k \)-th run. For each question \( q \), we compute adjacent prediction agreement over the sequence of repeated judgments. \textbf{Error Rate (ERR)} denotes the rate at which the LLM judge fails to correctly parse or follow the prescribed judgment format across different temperature settings. This metric indirectly captures the extent to which the judge model remains compliant with the evaluation instructions as temperature changes:
$$
\text{ERR} = \frac{1}{N}\sum_{i=1}^N \mathbf{1}(\text{format\_error}_i)
$$
Format errors catch unusable outputs that cannot be automatically parsed into clear winners. Pairwise needs ``A/B/C''; Single-answer judge type needs comparable [1,10] scores. 10\% of runs are not successfully parsed for clear winner or clear scoring.

\section{Results}

The experimental corpus comprises 500 base questions, evaluated under three judge
configurations, each operating with two prompting strategies---Direct base prompt and
CoT reasoning. Across six sampling temperatures, each question
is inferred 10 times per configuration using distinct random seeds while holding
all other generation hyperparameters constant. This large-scale experimental design not only enables robust empirical analysis but also provides sufficient statistical power to support rigorous causal inference. All results are available on Hugging Face for further exploration\footnote{\url{https://huggingface.co/datasets/Volavion/eval_temperatures_bench}}.

\subsection{Does temperature matter?}

\begin{table*}[htbp]
\centering
\begin{adjustbox}{width=1.00\textwidth}
\begin{tabular}{c>{\raggedright\arraybackslash}p{3.25cm}cccccccc}
\toprule
\multirow{2}{*}{\textbf{\faBalanceScale T \faBalanceScale}} & \multirow{2}{*}{\textbf{Model}} & \multirow{2}{*}{\textbf{Judge Type}} & \multirow{2}{*}{\textbf{\begin{tabular}[c]{@{}c@{}}Prompt\\ Strategy\end{tabular}}} & \multicolumn{3}{c}{\textbf{MT-Bench Human Judgments}} & \multicolumn{3}{c}{\textbf{MMLU\_Pro}} \\
\cmidrule(lr){5-7}\cmidrule(lr){8-10}
& & & & \textbf{\begin{tabular}[c]{@{}c@{}}AGR$\uparrow$ \\ ($\mu \pm$ std)\end{tabular}} & \textbf{\begin{tabular}[c]{@{}c@{}}CON$\uparrow$ \\ ($\mu \pm$ std)\end{tabular}} & \textbf{\begin{tabular}[c]{@{}c@{}}ERR$\downarrow$ \\ ($\mu \pm$ std)\end{tabular}} & \textbf{\begin{tabular}[c]{@{}c@{}}AGR$\uparrow$ \\ ($\mu \pm$ std)\end{tabular}} & \textbf{\begin{tabular}[c]{@{}c@{}}CON$\uparrow$ \\ ($\mu \pm$ std)\end{tabular}} & \textbf{\begin{tabular}[c]{@{}c@{}}ERR$\downarrow$ \\ ($\mu \pm$ std)\end{tabular}} \\
\midrule

\cellcolor{TColdBg} & \cellcolor{TColdBg} & \cellcolor{TColdBg} & \cellcolor{TColdBg}Direct & \cellcolor{TColdBg}\textbf{0.61 $\pm$ 0.00} & \cellcolor{TColdBg}0.98 $\pm$ 0.09 & \cellcolor{TColdBg}0.00 $\pm$ 0.00 & \cellcolor{TColdBg}0.58 $\pm$ 0.01 & \cellcolor{TColdBg}0.97 $\pm$ 0.12 & \cellcolor{TColdBg}0.00 $\pm$ 0.00 \\
\cellcolor{TColdBg} & \cellcolor{TColdBg} & \cellcolor{TColdBg}\multirow{-2}{*}{\begin{tabular}[c]{@{}c@{}}Pairwise\\Comparison\end{tabular}} & \cellcolor{TColdBg}CoT & \cellcolor{TColdBg}0.59 $\pm$ 0.01 & \cellcolor{TColdBg}0.94 $\pm$ 0.16 & \cellcolor{TColdBg}0.04 $\pm$ 0.01 & \cellcolor{TColdBg}0.55 $\pm$ 0.02 & \cellcolor{TColdBg}0.95 $\pm$ 0.16 & \cellcolor{TColdBg}0.25 $\pm$ 0.03 \\
\cellcolor{TColdBg} & \cellcolor{TColdBg} & \cellcolor{TColdBg} & \cellcolor{TColdBg}Direct & \cellcolor{TColdBg}0.54 $\pm$ 0.01 & \cellcolor{TColdBg}0.95 $\pm$ 0.14 & \cellcolor{TColdBg}0.00 $\pm$ 0.00 & \cellcolor{TColdBg}\textbf{0.70 $\pm$ 0.01} & \cellcolor{TColdBg}0.94 $\pm$ 0.16 & \cellcolor{TColdBg}0.00 $\pm$ 0.00 \\
\cellcolor{TColdBg} & \cellcolor{TColdBg} & \cellcolor{TColdBg}\multirow{-2}{*}{\begin{tabular}[c]{@{}c@{}}Single Answer\\Grading\end{tabular}} & \cellcolor{TColdBg}CoT & \cellcolor{TColdBg}0.54 $\pm$ 0.01 & \cellcolor{TColdBg}0.91 $\pm$ 0.18 & \cellcolor{TColdBg}0.01 $\pm$ 0.00 & \cellcolor{TColdBg}0.63 $\pm$ 0.02 & \cellcolor{TColdBg}0.83 $\pm$ 0.26 & \cellcolor{TColdBg}0.17 $\pm$ 0.02 \\
\cellcolor{TColdBg} & \cellcolor{TColdBg} & \cellcolor{TColdBg} & \cellcolor{TColdBg}Direct & \cellcolor{TColdBg}0.58 $\pm$ 0.01 & \cellcolor{TColdBg}0.94 $\pm$ 0.15 & \cellcolor{TColdBg}0.00 $\pm$ 0.00 & \cellcolor{TColdBg}0.65 $\pm$ 0.01 & \cellcolor{TColdBg}0.94 $\pm$ 0.16 & \cellcolor{TColdBg}0.00 $\pm$ 0.00 \\
\cellcolor{TColdBg} & \cellcolor{TColdBg}\multirow{-6}{*}{\begin{tabular}[c]{@{}l@{}}Qwen/Qwen3-Next\\80B-A3B-Instruct\\FP8\end{tabular}} & \cellcolor{TColdBg}\multirow{-2}{*}{\begin{tabular}[c]{@{}c@{}}Reference Guided\\Grading\end{tabular}} & \cellcolor{TColdBg}CoT & \cellcolor{TColdBg}0.58 $\pm$ 0.01 & \cellcolor{TColdBg}0.90 $\pm$ 0.20 & \cellcolor{TColdBg}0.05 $\pm$ 0.01 & \cellcolor{TColdBg}0.64 $\pm$ 0.02 & \cellcolor{TColdBg}0.86 $\pm$ 0.22 & \cellcolor{TColdBg}0.11 $\pm$ 0.01 \\
\cellcolor{TColdBg} & \cellcolor{TColdBg} & \cellcolor{TColdBg} & \cellcolor{TColdBg}Direct & \cellcolor{TColdBg}0.59 $\pm$ 0.00 & \cellcolor{TColdBg}\textbf{1.00 $\pm$ 0.04} & \cellcolor{TColdBg}0.00 $\pm$ 0.00 & \cellcolor{TColdBg}0.61 $\pm$ 0.00 & \cellcolor{TColdBg}\textbf{0.99 $\pm$ 0.08} & \cellcolor{TColdBg}0.00 $\pm$ 0.00 \\
\cellcolor{TColdBg} & \cellcolor{TColdBg} & \cellcolor{TColdBg}\multirow{-2}{*}{\begin{tabular}[c]{@{}c@{}}Pairwise\\Comparison\end{tabular}} & \cellcolor{TColdBg}CoT & \cellcolor{TColdBg}0.56 $\pm$ 0.00 & \cellcolor{TColdBg}0.99 $\pm$ 0.08 & \cellcolor{TColdBg}0.00 $\pm$ 0.00 & \cellcolor{TColdBg}0.53 $\pm$ 0.01 & \cellcolor{TColdBg}\textbf{0.98 $\pm$ 0.11} & \cellcolor{TColdBg}0.00 $\pm$ 0.00 \\
\cellcolor{TColdBg} & \cellcolor{TColdBg} & \cellcolor{TColdBg} & \cellcolor{TColdBg}Direct & \cellcolor{TColdBg}0.51 $\pm$ 0.00 & \cellcolor{TColdBg}0.98 $\pm$ 0.09 & \cellcolor{TColdBg}0.00 $\pm$ 0.00 & \cellcolor{TColdBg}0.56 $\pm$ 0.01 & \cellcolor{TColdBg}0.98 $\pm$ 0.11 & \cellcolor{TColdBg}0.00 $\pm$ 0.00 \\
\cellcolor{TColdBg} & \cellcolor{TColdBg} & \cellcolor{TColdBg}\multirow{-2}{*}{\begin{tabular}[c]{@{}c@{}}Single Answer\\Grading\end{tabular}} & \cellcolor{TColdBg}CoT & \cellcolor{TColdBg}0.48 $\pm$ 0.01 & \cellcolor{TColdBg}0.94 $\pm$ 0.15 & \cellcolor{TColdBg}0.00 $\pm$ 0.00 & \cellcolor{TColdBg}0.64 $\pm$ 0.01 & \cellcolor{TColdBg}0.91 $\pm$ 0.18 & \cellcolor{TColdBg}0.00 $\pm$ 0.00 \\
\cellcolor{TColdBg} & \cellcolor{TColdBg} & \cellcolor{TColdBg} & \cellcolor{TColdBg}Direct & \cellcolor{TColdBg}0.57 $\pm$ 0.00 & \cellcolor{TColdBg}\textbf{1.00 $\pm$ 0.04} & \cellcolor{TColdBg}0.00 $\pm$ 0.00 & \cellcolor{TColdBg}\textbf{0.69 $\pm$ 0.01} & \cellcolor{TColdBg}\textbf{0.99 $\pm$ 0.09 }& \cellcolor{TColdBg}0.00 $\pm$ 0.00 \\
\cellcolor{TColdBg}\multirow{-12}{*}{\textbf{\textcolor{TColdText}{\faSnowflake} 0.01 \textcolor{TColdText}{\faSnowflake}}} & \cellcolor{TColdBg}\multirow{-6}{*}{\begin{tabular}[c]{@{}l@{}}google/\\gemma-3-27b-it\end{tabular}} & \cellcolor{TColdBg}\multirow{-2}{*}{\begin{tabular}[c]{@{}c@{}}Reference Guided\\Grading\end{tabular}} & \cellcolor{TColdBg}CoT & \cellcolor{TColdBg}0.57 $\pm$ 0.01 & \cellcolor{TColdBg}0.96 $\pm$ 0.13 & \cellcolor{TColdBg}0.00 $\pm$ 0.00 & \cellcolor{TColdBg}0.68 $\pm$ 0.01 & \cellcolor{TColdBg}0.95 $\pm$ 0.15 & \cellcolor{TColdBg}0.00 $\pm$ 0.00 \\

\midrule

\cellcolor{THotBg} & \cellcolor{THotBg} & \cellcolor{THotBg} & \cellcolor{THotBg}Direct & \cellcolor{THotBg}0.59 $\pm$ 0.03 & \cellcolor{THotBg}0.84 $\pm$ 0.23 & \cellcolor{THotBg}0.07 $\pm$ 0.07 & \cellcolor{THotBg}0.54 $\pm$ 0.06 & \cellcolor{THotBg}0.68 $\pm$ 0.26 & \cellcolor{THotBg}0.17 $\pm$ 0.19 \\
\cellcolor{THotBg} & \cellcolor{THotBg} & \cellcolor{THotBg}\multirow{-2}{*}{\begin{tabular}[c]{@{}c@{}}Pairwise\\Comparison\end{tabular}} & \cellcolor{THotBg}CoT & \cellcolor{THotBg}0.53 $\pm$ 0.07 & \cellcolor{THotBg}0.66 $\pm$ 0.33 & \cellcolor{THotBg}0.37 $\pm$ 0.13 & \cellcolor{THotBg}0.45 $\pm$ 0.10 & \cellcolor{THotBg}0.61 $\pm$ 0.33 & \cellcolor{THotBg}\textbf{0.46 $\pm$ 0.15} \\
\cellcolor{THotBg} & \cellcolor{THotBg} & \cellcolor{THotBg} & \cellcolor{THotBg}Direct & \cellcolor{THotBg}0.52 $\pm$ 0.03 & \cellcolor{THotBg}0.76 $\pm$ 0.25 & \cellcolor{THotBg}0.00 $\pm$ 0.00 & \cellcolor{THotBg}0.60 $\pm$ 0.02 & \cellcolor{THotBg}0.72 $\pm$ 0.25 & \cellcolor{THotBg}0.00 $\pm$ 0.00 \\
\cellcolor{THotBg} & \cellcolor{THotBg} & \cellcolor{THotBg}\multirow{-2}{*}{\begin{tabular}[c]{@{}c@{}}Single Answer\\Grading\end{tabular}} & \cellcolor{THotBg}CoT & \cellcolor{THotBg}0.50 $\pm$ 0.02 & \cellcolor{THotBg}0.63 $\pm$ 0.29 & \cellcolor{THotBg}\textbf{0.13 $\pm$ 0.04} & \cellcolor{THotBg}0.51 $\pm$ 0.04 & \cellcolor{THotBg}0.58 $\pm$ 0.32 & \cellcolor{THotBg}0.35 $\pm$ 0.07 \\
\cellcolor{THotBg} & \cellcolor{THotBg} & \cellcolor{THotBg} & \cellcolor{THotBg}Direct & \cellcolor{THotBg}0.54 $\pm$ 0.06 & \cellcolor{THotBg}0.70 $\pm$ 0.26 & \cellcolor{THotBg}\textbf{0.13 $\pm$ 0.12} & \cellcolor{THotBg}0.60 $\pm$ 0.06 & \cellcolor{THotBg}0.71 $\pm$ 0.26 & \cellcolor{THotBg}0.17 $\pm$ 0.17 \\
\cellcolor{THotBg} & \cellcolor{THotBg}\multirow{-6}{*}{\begin{tabular}[c]{@{}l@{}}Qwen/Qwen3-Next\\80B-A3B-Instruct\\FP8\end{tabular}} & \cellcolor{THotBg}\multirow{-2}{*}{\begin{tabular}[c]{@{}c@{}}Reference Guided\\Grading\end{tabular}} & \cellcolor{THotBg}CoT & \cellcolor{THotBg}0.45 $\pm$ 0.06 & \cellcolor{THotBg}0.55 $\pm$ 0.30 & \cellcolor{THotBg}\textbf{0.37 $\pm$ 0.09} & \cellcolor{THotBg}0.44 $\pm$ 0.10 & \cellcolor{THotBg}0.57 $\pm$ 0.32 & \cellcolor{THotBg}\textbf{0.49 $\pm$ 0.08} \\
\cellcolor{THotBg} & \cellcolor{THotBg} & \cellcolor{THotBg} & \cellcolor{THotBg}Direct & \cellcolor{THotBg}0.58 $\pm$ 0.00 & \cellcolor{THotBg}0.93 $\pm$ 0.19 & \cellcolor{THotBg}0.00 $\pm$ 0.00 & \cellcolor{THotBg}0.59 $\pm$ 0.01 & \cellcolor{THotBg}0.90 $\pm$ 0.20 & \cellcolor{THotBg}0.00 $\pm$ 0.00 \\
\cellcolor{THotBg} & \cellcolor{THotBg} & \cellcolor{THotBg}\multirow{-2}{*}{\begin{tabular}[c]{@{}c@{}}Pairwise\\Comparison\end{tabular}} & \cellcolor{THotBg}CoT & \cellcolor{THotBg}0.55 $\pm$ 0.02 & \cellcolor{THotBg}0.86 $\pm$ 0.23 & \cellcolor{THotBg}0.07 $\pm$ 0.03 & \cellcolor{THotBg}0.53 $\pm$ 0.03 & \cellcolor{THotBg}0.75 $\pm$ 0.27 & \cellcolor{THotBg}0.09 $\pm$ 0.04 \\
\cellcolor{THotBg} & \cellcolor{THotBg} & \cellcolor{THotBg} & \cellcolor{THotBg}Direct & \cellcolor{THotBg}0.49 $\pm$ 0.01 & \cellcolor{THotBg}0.74 $\pm$ 0.25 & \cellcolor{THotBg}0.00 $\pm$ 0.00 & \cellcolor{THotBg}0.53 $\pm$ 0.05 & \cellcolor{THotBg}0.66 $\pm$ 0.24 & \cellcolor{THotBg}0.00 $\pm$ 0.00 \\
\cellcolor{THotBg} & \cellcolor{THotBg} & \cellcolor{THotBg}\multirow{-2}{*}{\begin{tabular}[c]{@{}c@{}}Single Answer\\Grading\end{tabular}} & \cellcolor{THotBg}CoT & \cellcolor{THotBg}0.47 $\pm$ 0.02 & \cellcolor{THotBg}0.63 $\pm$ 0.27 & \cellcolor{THotBg}0.01 $\pm$ 0.01 & \cellcolor{THotBg}0.56 $\pm$ 0.03 & \cellcolor{THotBg}0.57 $\pm$ 0.27 & \cellcolor{THotBg}0.03 $\pm$ 0.01 \\
\cellcolor{THotBg} & \cellcolor{THotBg} & \cellcolor{THotBg} & \cellcolor{THotBg}Direct & \cellcolor{THotBg}0.56 $\pm$ 0.01 & \cellcolor{THotBg}0.88 $\pm$ 0.21 & \cellcolor{THotBg}0.00 $\pm$ 0.00 & \cellcolor{THotBg}0.67 $\pm$ 0.03 & \cellcolor{THotBg}0.86 $\pm$ 0.25 & \cellcolor{THotBg}0.00 $\pm$ 0.00 \\
\cellcolor{THotBg}\multirow{-12}{*}{\textbf{\textcolor{THotText}{\faFire} 3.0 \textcolor{THotText}{\faFire}}} & \cellcolor{THotBg}\multirow{-6}{*}{\begin{tabular}[c]{@{}l@{}}google/\\gemma-3-27b-it\end{tabular}} & \cellcolor{THotBg}\multirow{-2}{*}{\begin{tabular}[c]{@{}c@{}}Reference Guided\\Grading\end{tabular}} & \cellcolor{THotBg}CoT & \cellcolor{THotBg}0.55 $\pm$ 0.02 & \cellcolor{THotBg}0.73 $\pm$ 0.27 & \cellcolor{THotBg}0.10 $\pm$ 0.02 & \cellcolor{THotBg}0.65 $\pm$ 0.04 & \cellcolor{THotBg}0.66 $\pm$ 0.29 & \cellcolor{THotBg}0.10 $\pm$ 0.04 \\
\bottomrule
\end{tabular}
\end{adjustbox}
\caption{Quantitative assessment of LLM-as-a-Judge stability across three sampling temperatures ($T \in \{0.01, 3.0\}$), two judge models, three evaluation paradigms (Pairwise Comparison, Single Answer Grading, and Reference-Guided Grading), and two prompting strategies (Direct and CoT), measured by agreement, consistency, and error rate on MT-Bench and MMLU-Pro. \textbf{Bolded} values indicate the highest and second-highest values globally within each metric--dataset combination.}
\label{tab:main_results}
\end{table*}

As shown in Table~\ref{tab:main_results}, performance is reported at temperatures 0.01, 1.00, and 3.00. At low temperature, all configurations exhibit exceptionally high consistency and near-zero error rates, indicating that LLM-judge outputs are highly stable and reproducible. At high temperature, consistency degrades substantially and error rates rise markedly, showing that high temperatures severely compromise judge reliability. In contrast, \texttt{gemma-3-27b-it} remains comparatively robust at high temperatures, maintaining higher consistency and lower error rates than \texttt{Qwen3-Next-80B}. Notably, for \texttt{Qwen3-Next-80B} under pairwise CoT evaluation, performance at $T{=}1.0$ is comparable to that at $T{=}0.01$: despite a higher error rate, the drop in \textit{AGR} is marginal. This suggests that moderate temperatures may improve alignment, highlighting the value of multiple trials with different random seeds. \textit{Ensemble judging}---aggregating multiple high-temperature samples via majority voting---may thus prove more robust than a single low-temperature pass.

\begin{figure}[htbp]
    \centering
    \includegraphics[width=1.0\linewidth]{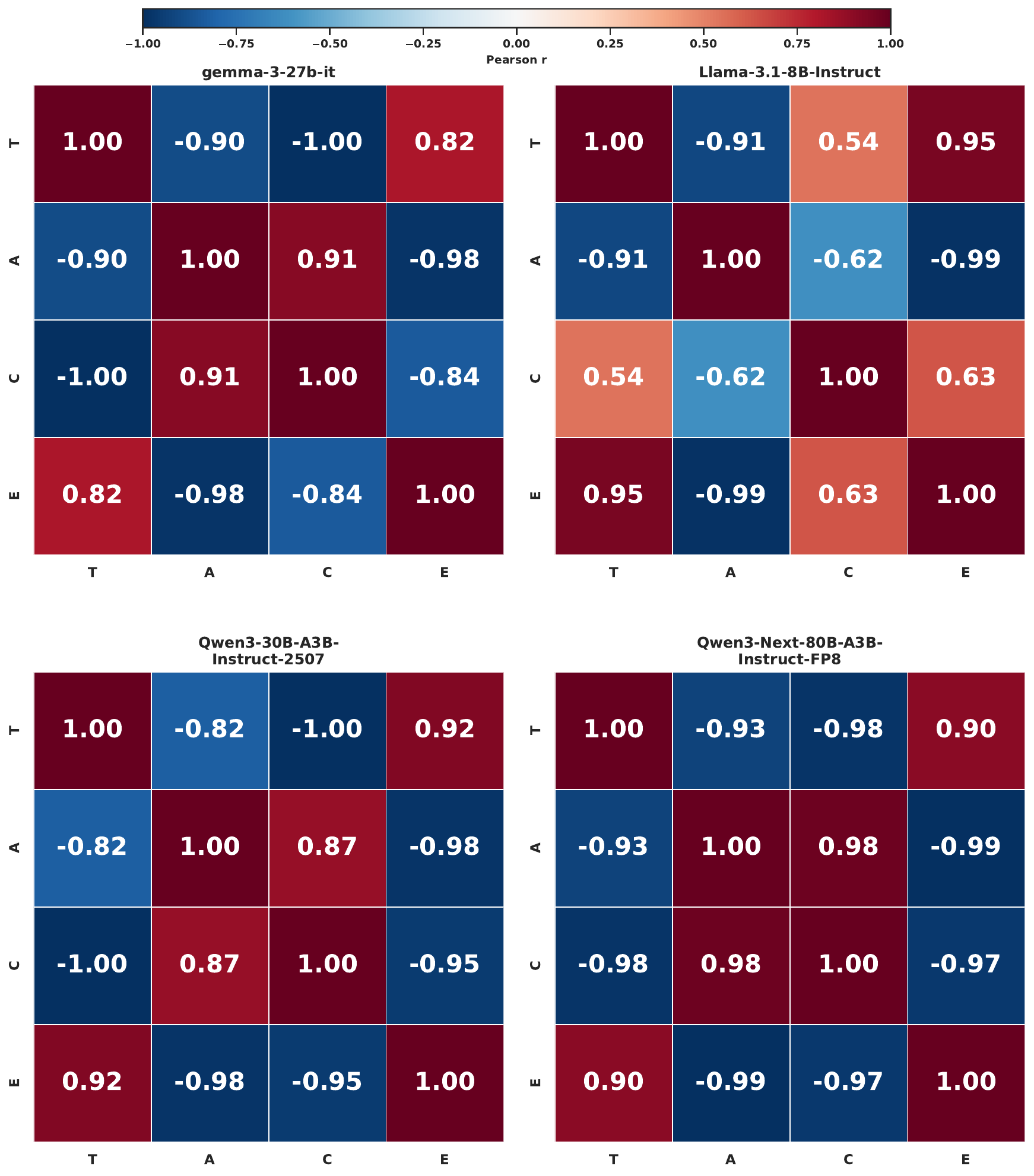} 
    \caption{Pearson correlation matrices between decoding temperature (T) and evaluation metrics—agreement (A), consistency (C), and error rate (E)—across four LLMs. We additionally report statistically significant results for Qwen3-30B-A3B-Instruct-2507 (Qwen3-30B) and Llama-3.1-8B-Instruct.}
    \label{fig:correlation_analysis}
\end{figure}

Finally, in the context of LLM-as-a-Judge, the degradation of 
\textit{Error Rate} and \textit{Consistency} at high temperatures is 
itself a valuable diagnostic signal, as it exposes the fragility 
boundaries of the model Judge.  At low temperatures, the standard deviation is minimal; at high 
temperatures, it increases sharply, indicating that Judge outcomes 
for the same input become highly unstable.
The standard deviation of \texttt{gemma-3-27b-it} remains relatively 
small at high temperatures, whereas that of \texttt{Qwen3-Next-80B} 
is substantially larger. This reveals a structural difference in the two models' sensitivity 
to temperature perturbation---a finding that warrants further 
investigation. It also implies that \textbf{model selection} should be informed by 
model-specific characteristics, as well as by whether output 
diversity is desired.

As shown in Figure~\ref{fig:correlation_analysis}, increasing temperature \(T\) consistently decreases Agreement \(A\) and markedly increases Error \(E\), indicating reduced evaluation reliability at higher temperatures. For \textit{Gemma-27b}, \textit{Qwen3-30B}, and \textit{Qwen3-Next-80B}, we observe near-perfect negative correlations between \(T\) and Consistency \(C\), as well as between \(A\) and \(E\), reflecting highly stable and consistent behavior across models. In contrast, \textit{Llama-3.1-8B-Instruct} deviates from this pattern, showing a positive correlation between \(T\) and \(C\), and between \(C\) and \(E\). This suggests that, as a smaller model, its judgments approach random guessing and are more sensitive to variations in random seeds.






\begin{figure*}[htbp]
  \centering
  \includegraphics[width=\linewidth]{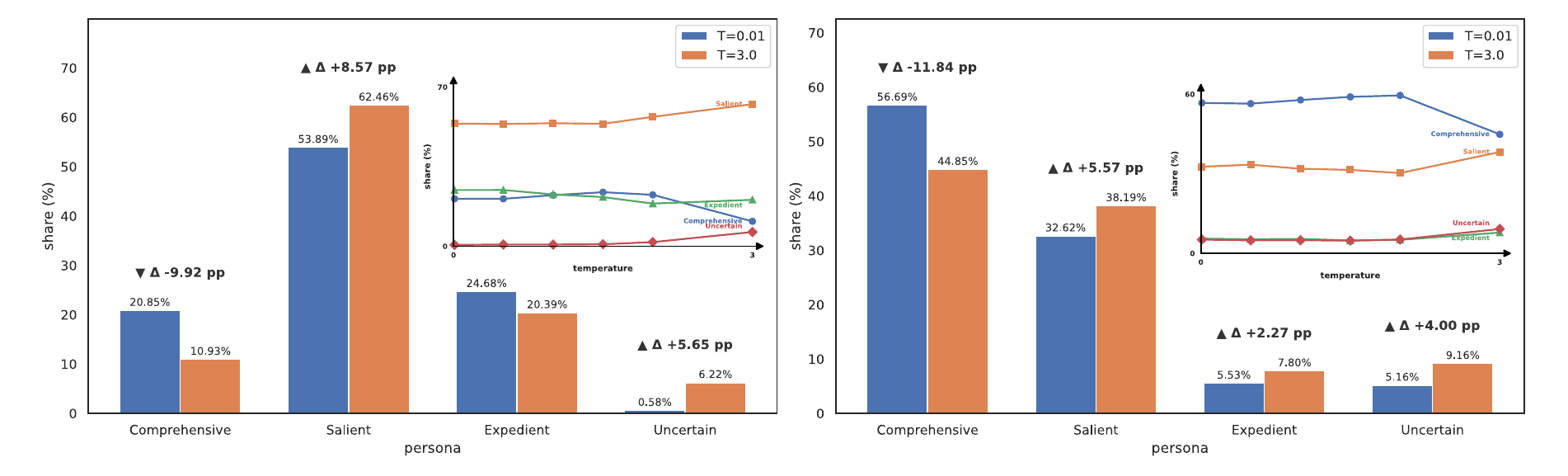}
  \caption{Temperature's effect on LLM-judge persona distributions. Left: gemma-3-27B; right: Qwen3-Next-80B. Bar charts compare persona compositions at the lowest and highest temperatures; line charts trace the four personas across the full temperature range.}
  \label{fig:persona-vs-temp}
\end{figure*}

\subsection{Judge Behaviors}
\paragraph{Temperature on Judge Personas} 
Taken together, Figure \ref{fig:persona-vs-temp} shows that temperature does not linearly enhance the comprehensiveness of judge reasoning; rather, it acts as a perturbation that reshapes the judge's existing reasoning persona. From low to moderate temperatures, the distribution of personas remains broadly stable, with pronounced shifts emerging only at higher temperatures—suggesting a threshold effect in persona transition. This transition also varies markedly across models: for Qwen, high temperatures disperse an otherwise concentrated judging pattern, yielding more diverse reasoning styles; for Gemma, high temperatures instead push the judge toward salient-cue-dominated and uncertain judging modes. High temperature is therefore best understood as increasing the plasticity of the judge's persona: it can introduce complementary judging perspectives, but may also expose the model's reliance on specific cues and its behavioral instability. For more details on the response, see Appendix~\ref{sec:high vs low}

\paragraph{Confusion at Low Temperatures}  A case-by-case inspection reveals that at higher temperatures, the model’s ability to adhere to the required JSON format degrades substantially, consistent with the observed increase in error rates: the model may omit the judge’s reasoning or fail to produce valid JSON outputs. Unexpectedly, even at low temperatures (\(T = 0.01\)), we still observe a non-negligible rate of formatting violations. This suggests that output randomness in LLMs is not fully eliminated under low-temperature decoding. These findings indicate that, in LLM-as-a-Judge settings, varying random seeds and temperature values can be useful for exposing variability and obtaining more robust evaluation signals.

\paragraph{Naturalistic Expression} At higher temperatures, the judge begins to sound less like a rule-follower and more like a person thinking out loud. Its outputs become more varied, fragmented, and spontaneous qualities largely absent at low temperatures, where decoding is more deterministic and strictly instruction-following. As temperature rises, the model follows the prompt flexibly, with a less rigid conversational style and starts to explore, producing responses that feel more natural. For qualitative examples, see Appendix~\ref{sec:llm_response_different_temp}.

\begin{tcolorbox}[
  enhanced,
  colback=teal!5,
  colframe=teal!60,
  boxrule=0.8pt,
  arc=3mm,
  left=8pt, right=8pt, top=6pt, bottom=6pt,
  title=\textcolor{teal!80!black}{\textit{Sample judge response at high temperature}},
  fonttitle=\small\bfseries,
  coltitle=teal!80!black,
  attach boxed title to top left={yshift=-2mm, xshift=4mm},
  boxed title style={colback=teal!10, colframe=teal!60, boxrule=0.5pt, arc=1mm}
]
\textcolor{teal!80!black}{\textbf{Assistant A (correct):}} correctly assigned ratings based on each review's sentiment. Review~1 (``phenomenal'', 5-point scale) $\rightarrow$ \textcolor{green!60!black}{\textbf{5}}; Review~2 (``unfortunately poor and predictable'', described as ``worst'') $\rightarrow$ \textcolor{green!60!black}{\textbf{1}}; Revision of ``okay'', ``forgettable'' $\rightarrow$ \textcolor{green!60!black}{\textbf{2}}, which corresponds to a midrange score (around 1.5--2.5 out of 5), similar to the ``middle ground'' rating of 3.

\vspace{4pt}
\textcolor{orange!80!black}{\textbf{Assistant B (incorrect):}} misassigned the scores. \emph{\textcolor{orange!80!black}{Review~A, clearly very positive, deserves a 5, while ``worst... in 2022'' is a negative review (hence 1).}} However, Assistant~B gave Review~2 a score of \textcolor{red!70!black}{\textbf{3}} (average or neutral?), even though it was clearly negative. Review~3 described the work as mediocre---thus a middle-ground rating around 2 or 3 would be appropriate. Instead, Assistant~B assigned it a \textcolor{red!70!black}{\textbf{1}}, the opposite of what was written.
\end{tcolorbox}

\subsection{Thermo-judge}

As shown in the previous section, LLMs can exhibit distinct \textit{judge personas} under different temperature settings. To exploit this, we propose a simple approach, \textbf{Thermo-Judge}: given an input query $x$, the judge model is run $N$ times across one or more temperature configurations, producing a set of verdicts; majority voting then selects the most frequent verdict as the final decision. We evaluate this method on three definitions of \textit{ambiguous cases} —those where the model struggles to reach a verdict—allowing us to test its robustness across different categorizations.

\paragraph{Def.~1 (Temp-Sensitive).}
A sample is \textbf{Ambiguous} if it shows both low-temperature consistency (Shannon entropy of verdicts over 10 repetitions at $T{=}0.01$ below $0.01$ bits) and high-temperature disagreement (average entropy over $T \in \{1.5, 2.0, 3.0\}$ above $0.50$ bits). This yields an Ambiguous subset of 42.6\%; all remaining samples are Non-ambiguous.

\paragraph{Def.~2 (Heuristic).}
To avoid circularity from temperature-sensitive signals, we define Ambiguous samples using external cues only: a human-annotated tie, high lexical overlap between candidates, or similar response lengths. About 36\% of samples fall into this subset.

\paragraph{Def.~3 (LLM-Labeled).}
As an orthogonal check, a frontier model (GPT-5.4) acts as an independent meta-judge, labeling each sample as ambiguous or not. Since this judgment is decoupled from the evaluated judge's decoding temperature, it provides a temperature-independent notion of ambiguity.

We evaluate a set of ensemble configurations that vary temperature \(T\) and sample size \(N\): \textbf{Ens-Low} (\(T \in \{0.01, 0.5, 1.0\}\), \(N{=}10\)) and \textbf{Ens-High} (\(T \in \{1.5, 2.0, 3.0\}\), \(N{=}10\) per temperature), where \textbf{Ens} (ensemble) aggregates \(N\) independent generations via majority voting. We further include \textbf{Ens-Anneal}, a token-level temperature annealing variant that samples the reasoning span at a high temperature \(T_{\text{reason}} \in \{2, 3\}\) and switches to a low temperature \(T_{\text{result}} \in \{0.01, 0.5\}\) upon the verdict trigger, combining exploratory diversity with decisive convergence. By coupling temperature-induced diversity with majority voting, these ensembles yield more robust and reliable judgments, particularly on ambiguous cases.

\begin{table}[t]
\centering
\small
\begin{adjustbox}{width=\columnwidth,center}
\begin{tabular}{l c c c c}
\toprule
Subset & \(T\) & \textit{Mean Entropy} & \textit{Flip Rate} & \textit{Lift} \\
\toprule
\rowcolor{gray!5}
\multirow{6}{*}{\textbf{Ambiguous}}
& 0.01 & 0.0000 & 0.0000 & 0.0000 \\
& 0.50 & 0.1722 & 0.0323 & 0.1722 \\
& 1.00 & 0.3298 & 0.0484 & 0.3298 \\
& 1.50 & 0.5720 & 0.0710 & 0.5720 \\
& 2.00 & 0.8068 & 0.1161 & 0.8068 \\
& 3.00 & \textbf{1.1003} & \textbf{0.2968} & \textbf{1.1003} \\
\midrule
\rowcolor{gray!8}
\multirow{6}{*}{\textbf{Non-Ambiguous}}
& 0.01 & 0.0517 & 0.0000 & 0.0000 \\
& 0.50 & 0.0540 & 0.0190 & 0.0022 \\
& 1.00 & 0.0630 & 0.0214 & 0.0112 \\
& 1.50 & 0.0681 & 0.0196 & 0.0165 \\
& 2.00 & 0.0979 & 0.0268 & 0.0462 \\
& 3.00 & \textbf{0.2709} & \textbf{0.0410} & \textbf{0.2192} \\
\bottomrule
\end{tabular}
\end{adjustbox}
\caption{Performance across temperatures for Ambiguous and Non-Ambiguous subsets. \textit{Mean Entropy}: average Shannon entropy of verdicts; \textit{Flip Rate}: fraction of majority changes w.r.t. \(T{=}0.01\); \textit{Lift}: entropy increase over \(T{=}0.01\). Bold indicates the maximum within each subset and metric.}
\label{tab:exp1f_single_max_per_group}
\end{table}

\begin{table*}[htbp]
\small
\centering
\begin{minipage}[t]{0.75\textwidth}
\vspace{0pt}
\begin{adjustbox}{width=\textwidth}
\begin{tabular}{l cc cc cc}
\toprule
& \multicolumn{2}{c}{\textbf{Temp-Sensitive}}
& \multicolumn{2}{c}{\textbf{Heuristic}}
& \multicolumn{2}{c}{\textbf{LLM-Labeled}} \\
\cmidrule(lr){2-3} \cmidrule(lr){4-5} \cmidrule(lr){6-7}
& Amb & Non-Amb & Amb & Non-Amb & Amb & Non-Amb \\
\midrule
\rowcolor{gray!5}
\multicolumn{7}{l}{\textbf{Chain-of-Thought (gemma-3-27B)}} \\
\addlinespace[2pt]
Ens-Low (T=0.01, N=10)         & 0.4460 & 0.6307 & 0.2549 & 0.6830 & 0.3708 & 0.7198 \\
Ens-Low (T=0.5, N=10)          & 0.4272 & 0.6376 & 0.2353 & 0.6859 & 0.3667 & 0.7160 \\
Ens-Low (T=1.0, N=10)          & 0.4413 & \textbf{0.6446} & 0.2614 & 0.6888 & 0.3792 & 0.7237 \\
Ens-High (T=1.5, N=10)         & 0.4507 & \textbf{0.6446} & 0.2680 & \textbf{0.6916} & 0.3750 & \textbf{0.7354} \\
Ens-High (T=2.0, N=10)         & 0.4460 & 0.6341 & 0.2484 & 0.6888 & 0.3667 & 0.7276 \\
Ens-High (T=3.0, N=10)         & 0.4648 & 0.6376 & 0.2876 & 0.6859 & \textbf{0.3917} & 0.7237 \\
\midrule
\rowcolor{gray!5}
\multicolumn{7}{l}{\textbf{Token-Level Temperature Annealing ($T_{\text{reason}} \rightarrow T_{\text{result}}$)}} \\
\addlinespace[2pt]
Ens-Anneal (2$\rightarrow$0.01, N=10)      & 0.4460 & 0.6411 & 0.2549 & \textbf{0.6916} & 0.3750 & 0.7276 \\
Ens-Anneal (2$\rightarrow$0.5, N=10)       & 0.4507 & 0.6411 & 0.2614 & \textbf{0.6916} & 0.3750 & 0.7315 \\
Ens-Anneal (3$\rightarrow$0.01, N=10)      & \textbf{0.4789} & 0.6307 & \textbf{0.2941} & 0.6859 & \textbf{0.3917} & 0.7276 \\
Ens-Anneal (3$\rightarrow$0.5, N=10)       & 0.4648 & 0.6341 & 0.2745 & 0.6888 & 0.3875 & 0.7237 \\
\bottomrule
\end{tabular}
\end{adjustbox}
\end{minipage}\hfill
\begin{minipage}[t]{0.23\textwidth}
\vspace{0pt}
\centering
\includegraphics[width=\linewidth]{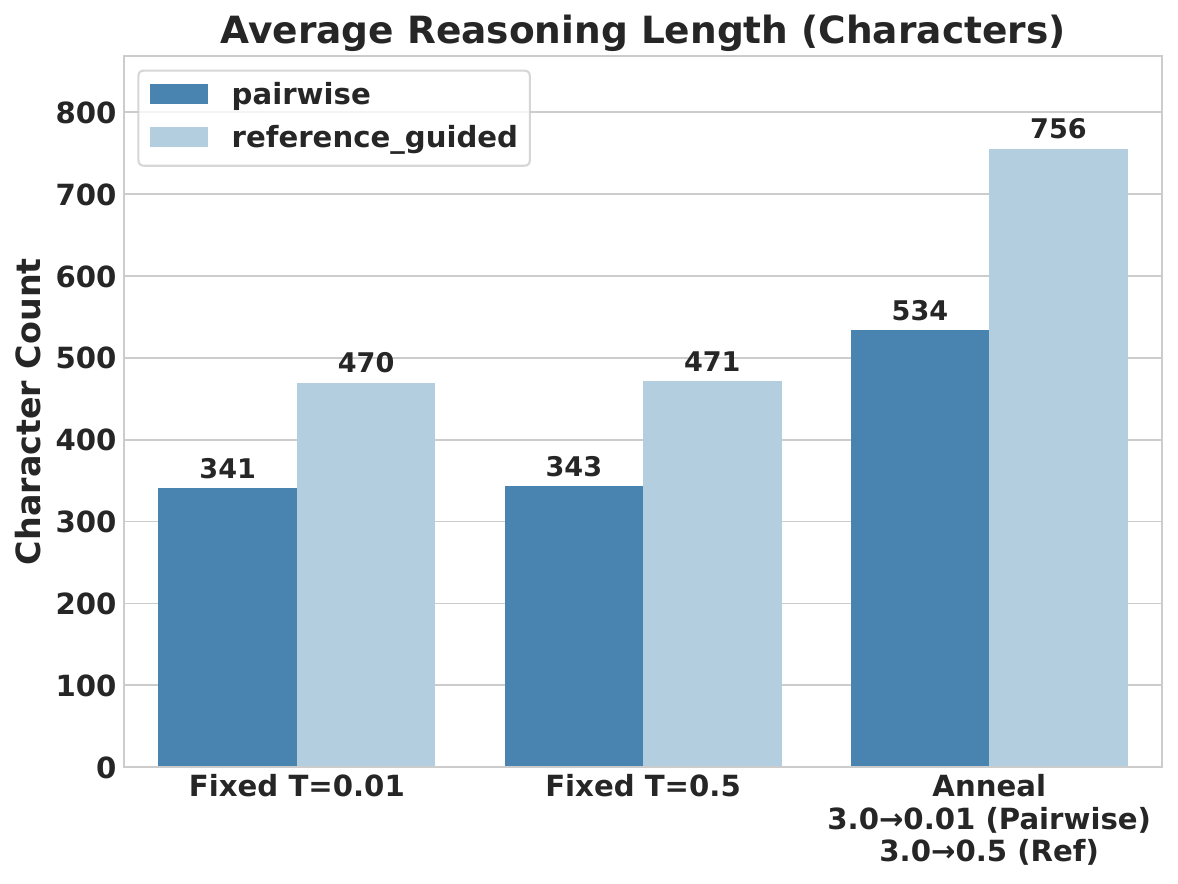}\\[0.6ex]
\includegraphics[width=\linewidth]{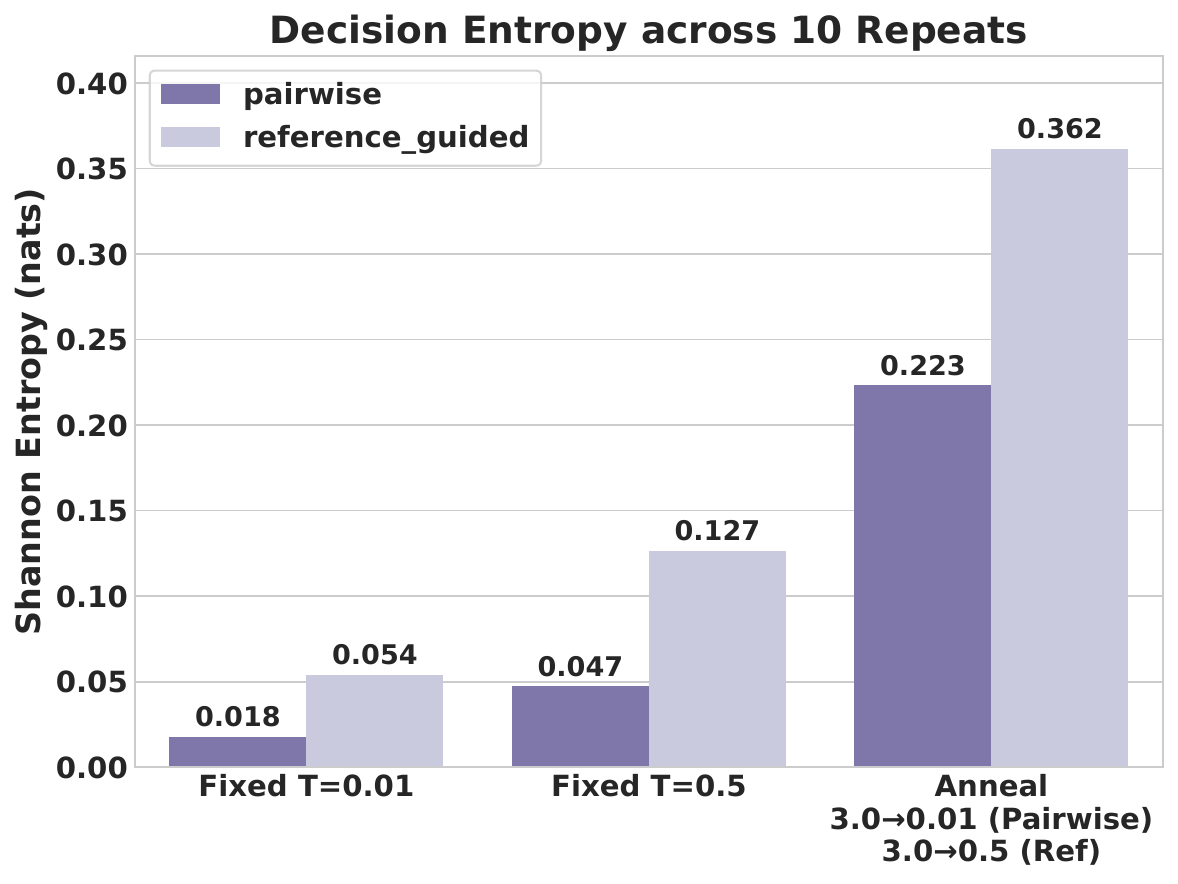}
\end{minipage}
\caption{Token-level temperature annealing for CoT judging on gemma-3-27B under three ambiguity definitions. We report Gemma-3-27B as a representative case; other evaluated models exhibit qualitatively similar trends. Bold marks the best per column; high-temperature ensembles lead on ambiguous cases, while low temperatures favor non-ambiguous ones. Right: average output length (top) and decision entropy over 10 runs, averaged across 500 questions (bottom), comparing fixed baselines ($T{=}0.01$, $0.5$) with the best annealing setup per task ($3.0{\rightarrow}0.01$ pairwise; $3.0{\rightarrow}0.5$ reference-guided).}
\label{tab:exp2_baseline_cot_three_defs}
\end{table*}

\subsubsection{Ambiguity Indicator}

As shown in Table~\ref{tab:exp1f_single_max_per_group}, ambiguous samples are markedly more sensitive to temperature: verdict entropy rises substantially, flip rates climb sharply, and disagreement lift grows monotonically with temperature. Non-ambiguous samples, by contrast, remain largely stable across all metrics, exhibiting only marginal changes in entropy, flip rate, and consistency. Low-temperature decoding therefore induces a form of \textbf{superficial certainty}: judgments may appear highly consistent while not necessarily being reliable, as the limited output variability constrains the judge's capacity for comprehensive, multi-perspective reasoning. Higher temperatures, in turn, expose the underlying uncertainty of ambiguous cases while leaving clear cases largely unaffected. Notably, because ambiguous cases respond strongly to temperature while clear cases do not, the degree of verdict flipping under repeated high-temperature sampling can serve as a practical indicator of case difficulty.

\subsubsection{Increasing the Temperature}

Under all three ambiguity definitions, agreement with human judgments on ambiguous cases generally rises with the reasoning-stage temperature, though not strictly monotonically. The best configuration is annealing with high-temperature reasoning and near-greedy verdict decoding ($3{\rightarrow}0.01$), which outperforms low-temperature ensembling ($T{=}0.01$) by roughly 3.3, 3.9, and 2.1 percentage points across the three definitions. Fixed high-temperature ensembling ($T{=}3.0$) follows closely, while low-temperature ensembles such as $T{=}0.5$ perform worst. This suggests that ambiguous cases benefit from multiple plausible judging perspectives: high-temperature reasoning lets the model explore diverse argumentative paths, and majority voting then converges toward human-like judgments. The right-hand panels corroborate this: annealing increases output length by roughly 60\% and markedly raises decision entropy across the 10 repeats (from about 0.02 to 0.22 for pairwise), making the ensemble meaningful. In contrast, at low temperatures the outputs are nearly identical across repeats, so ensembling effectively degenerates into a single judgment. Moreover, holding the reasoning temperature fixed at 3, lowering the verdict temperature from 0.5 to 0.01 yields a slight further gain, supporting the division of labor between ``exploratory reasoning and decisive verdicts.''

On non-ambiguous cases, the gap across configurations remains within roughly 2 percentage points. Neither high-temperature reasoning nor annealing degrades performance, and low-temperature settings show no clear advantage. The gains from annealing on ambiguous cases therefore do not come at the expense of easy cases. High-temperature reasoning paired with low-temperature verdict decoding thus constitutes a decoding strategy that is consistently effective on ambiguous cases while remaining side-effect-free on non-ambiguous ones.

\section{Conclusion}
This work presents a large-scale, controlled study of how decoding temperature shapes both the reliability and reasoning behavior of LLM judges, offering actionable guidance for practice and future research. Low temperatures yield near-perfect consistency and negligible error rates, whereas higher temperatures sharply degrade consistency and inflate error rates, with effects varying by model scale, prompting strategy, and evaluation paradigm. Beyond reliability, temperature reshapes how judges reason: we characterize this through four judge personas---from the quick, narrow Expedient Judge to the broad but hedging Uncertain Judge---whose distributions shift systematically with temperature. High temperatures elicit more exploratory, multi-perspective reasoning, albeit at the cost of occasional instruction violations and parsing errors; notably, even low temperatures do not fully eliminate such violations, challenging the assumption that low temperature is universally optimal.

Crucially, temperature has a higher effect on ambiguous cases: clear ones stay stable, while ambiguous ones flip often under repeated high-temperature sampling---a built-in difficulty indicator. \textbf{Thermo-Judge} can exploit this by pairing high-temperature exploration with low-temperature, majority-voted arbitration, gaining persona diversity while curbing heat-induced errors. It improves agreement on ambiguous cases without affecting the clear ones.

\section{Limitations}

We outline several limitations of this study. First, we restrict our experiments to open-source models. This choice is primarily driven by the need for transparent and controllable temperature configurations, which are not consistently available in closed-source systems. While this may limit direct coverage of proprietary models, the controlled setting allows us to isolate and systematically analyze temperature effects, and we expect the qualitative trends to generalize across model families.

Second, inference outcomes may vary with random seeds, temperature settings, and hardware environments. This effect becomes more pronounced at higher temperatures (up to \(T = 3.0\) in our experiments), where generation is inherently more stochastic. Even with identical seeds and decoding parameters, minor differences in floating-point computations across GPU architectures can occasionally lead to divergent outputs. Importantly, our results are based on aggregate statistics over multiple runs, which mitigates the impact of such variability. Exploring even higher temperature regimes remains an interesting direction for future work.

For future work, a promising direction for improving judgment quality—particularly in agent-based systems—is to combine a high-temperature agent for exploratory initial judgments with a low-temperature agent for output refinement, thereby balancing reasoning diversity with determinism. More broadly, these findings advocate for task-adaptive temperature strategies in LLM evaluation pipelines, moving beyond fixed low-temperature defaults to enhance robustness. We do not explore the interaction between temperature and various evaluator biases, which remains an open research question.

\bibliography{custom}

\appendix

\section{Prompt For Generating Gold Judge Labels}
\label{appendix:judge_prompt_4.5}

To evaluate the MMLU-Pro dataset in the absence of human\-judges results, we employ the state-of-the-art (SOTA) GPT-5.4 model as an automated judge. 
The evaluation protocol, illustrated in the following prompt box, determines the superior response between two model outputs or declares a tie when performance is deemed equivalent.

\begin{promptbox}{System Prompt Template}
You are an impartial judge evaluating answers to multiple-choice questions.  
Your task is to decide which answer is better or whether they are equally good.

Respond \textbf{ONLY} with a JSON object — no markdown, no extra text — in this exact format:

\{"winner": "<A|B|tie>", "reason": "<one concise sentence>"\}
\end{promptbox}

\begin{userbox}{User Prompt Template}
\textbf{Question:}
\texttt{[question text]}

\textbf{Options:}
\texttt{A. [option A]}
\texttt{B. [option B]}
\texttt{...}

\textbf{Correct Answer:}
\texttt{[correct answer]}

\textbf{Answer A (model: [modelA]):}
\texttt{[answer A content]}

\textbf{Answer B (model: [modelB]):}
\texttt{[answer B content]}

Which answer is better? Reply with JSON only.
\end{userbox}

\section{Prompt Template For Judge}
\label{appendix:prompt template}

\begin{userbox}{Pairwise Comparison Direct Judge Prompt}
You are a fair and impartial judge. Your task is to evaluate two AI assistant responses to a user question and decide which one is better.

Respond ONLY with valid JSON using exactly one key:
\begin{verbatim}
{"judge result":"A"}
\end{verbatim}

Use ``A'' if assistant A is better, ``B'' if assistant B is better, and ``C'' for a tie.
Do not output any additional keys, explanation, markdown, or text.
\end{userbox}

\begin{userbox}{Pairwise Comparison COT Judge Prompt}
You are a fair and impartial judge. Your task is to evaluate two AI assistant responses to a user question. First, compare them on helpfulness, relevance, accuracy, depth, and creativity. Then compare them on coherence, clarity, and language quality. Finally, give your verdict.

Respond ONLY with valid JSON using exactly these two keys:
\begin{verbatim}
{"judge reason":"<brief explanation>"
,"judge result":"A"}
\end{verbatim}

Use ``A'' if assistant A is better, ``B'' if assistant B is better, and ``C'' for a tie.
Do not output markdown or any extra text outside the JSON object.
\end{userbox}

\begin{userbox}{Reference Guided Grading Direct Judge Prompt}
Please act as an impartial judge and evaluate the quality of the responses provided by two AI assistants to the user question displayed below. Your evaluation should consider correctness and helpfulness. You will be given a reference answer, assistant A's answer, and assistant B's answer. Your job is to evaluate which assistant's answer is better. Begin by comparing both assistants' answers with the reference answer. Identify mistakes if needed. Avoid position bias, length bias, and any bias toward assistant names.

Respond ONLY with valid JSON using exactly one key:
\begin{verbatim}
{"judge result":"A"}
\end{verbatim}

Use ``A'' if assistant A is better, ``B'' if assistant B is better, and ``C'' for a tie.
Do not output any additional keys, explanation, markdown, or text.
\end{userbox}

\begin{userbox}{Reference Guided Grading COT Judge Prompt}
Please act as an impartial judge and evaluate the quality of the responses provided by two AI assistants to the user question displayed below. Your evaluation should consider correctness and helpfulness. You will be given a reference answer, assistant A's answer, and assistant B's answer. Your job is to evaluate which assistant's answer is better.

First, compare assistant A's answer and assistant B's answer against the reference answer for correctness, completeness, and helpfulness. Then identify any mistakes, omissions, or misleading claims in each answer. Finally, compare the two assistants directly and explain which answer is better overall, while avoiding position bias, length bias, and any bias toward model names.

Respond ONLY with valid JSON using exactly these two keys:
\begin{verbatim}
{"judge_reason":"<brief explanation>"
,"judge result":"A"}
\end{verbatim}

Use ``A'' if assistant A is better, ``B'' if assistant B is better, and ``C'' for a tie.
Do not output markdown or any extra text outside the JSON object.
\end{userbox}

\begin{userbox}{Single Answer Grading Direct Judge Prompt}
You are a fair and impartial judge. Rate the quality of the AI assistant's response to the user question on a scale of 1 to 10.

Respond ONLY with valid JSON using exactly one key:
\begin{verbatim}
{"judge result":7}
\end{verbatim}

The value of \texttt{judge\_result} must be an integer between 1 and 10.
Do not output any additional keys, explanation, markdown, or text.
\end{userbox}

\begin{userbox}{Single Answer Grading COT Judge Prompt}
You are a fair and impartial judge. Rate the quality of the AI assistant's response to the user question on a scale of 1 to 10. First analyze the response's helpfulness, relevance, accuracy, depth, creativity, and coherence. Then provide your rating.

Respond ONLY with valid JSON using exactly these two keys:
\begin{verbatim}
{"judge_reason":"<brief explanation>"
,"judge result":7}
\end{verbatim}

The value of \texttt{judge\_result} must be an integer between 1 and 10.
Do not output markdown or any extra text outside the JSON object.
\end{userbox}

\section{Examples}
\subsection{High Temperature Vs Low Temperature}
\label{sec:high vs low}

This response from Qwen3-30B-A3B-Instruct-2507 is more direct and more polite, yet considers fewer aspects and yields a more concise answer from the Judge.
\begin{responsebox}{LLM Response With T=0.01}

Assistant A's answer is incorrect. The probability is not 1/3. The correct approach is to calculate
the probability of the complementary event (rolling a total less than 3) and subtract it from 1. The
only outcome with a total less than 3 is (1,1), which gives a sum of 2. So the probability of
rolling a total of at least 3 is $1 - 1/36 = 35/36$. Assistant B correctly identifies all 36 possible
outcomes and correctly notes that only (1,1) results in a sum less than 3. However, Assistant B
makes a critical error by stating that all 36 outcomes result in a total of at least 3, which is
false. The outcome (1,1) gives a sum of 2, which is less than 3. Therefore, there are actually 35
favorable outcomes, not 36. The correct probability is $35/36$, not 1. Both assistants are incorrect,
but Assistant B provides a more thorough and correct reasoning process, even though the final
conclusion is wrong. Assistant A's answer is completely wrong in both reasoning and final answer.
Therefore, Assistant B is better despite the error in the final conclusion.
\end{responsebox}

This response, in contrast, provides more detailed information and a greater number of justifications for their judgment. When the temperature is set to a high value, the model does not produce nonsensical outputs or exhibit obvious hallucinations; instead, it offers a more detailed explanation of its analysis to make its decisions appear well grounded. Furthermore, the final conclusion reached is correct, which is highly significant because it reflects the meaningful role of the high‑temperature setting.

\begin{responsebox}{LLM Response With T=2.0}
Assistant A's answer contains a significant error: claiming the probability is 1/3 (33.3\%) is
incorrect. The only possible totals that are less than 3 are when rolling a 2. There is only one
outcome that produces a sum of 2: (1,1). Thus, the probability of rolling at least 3 is $35/36$, not
1/3. Additionally, Assistant A's justification (``probability of rolling a 3 or higher is 1/3 of the
total number of outcomes (6)'') is illogical and mathematically incorrect---total possible outcomes for
the sum are not 6, nor does the number 6 relate to the probability in the claimed manner.

Assistant B provides a comprehensive, accurate calculation using the sample space of all 36 possible
outcomes, correctly identifying that only (1,1) produces a sum less than 3, hence the probability of
rolling at least 3 is $35/36$. However, Assistant B incorrectly asserts that `there are 36 such
outcomes' and therefore the probability is $36/36=1$---an erroneous conclusion from overcounting. The
correct number of favorable outcomes (for sum $\geq$ 3) should be 35, not 36.

Although both answers contain calculation errors, Assistant B's reasoning process is vastly more
accurate in approach and method. It correctly lists and counts the full set of sample space
elements. The error in Assistant B's final tally (missing that (1,1) has sum of 2) seems to be a
lapse, but does not undermine the overall clarity and correctness of the reasoning path and
approach.

In comparison, Assistant A provides no such method, misidentifies the distribution completely, and
gives an obviously wrong probabilistic fraction.

Thus, B provides more helpful, relevant, and accurate reasoning and methodology---even if there is a
minor factual mistake in conclusion---making it vastly better overall than A.
\end{responsebox}

For the same question, this response is likewise reasonable with T=3.0. In my view, the Judge’s reasoning is more comprehensive, and its final decision of a tie suggests that it detected certain flaws in both responses, making it difficult to determine a clear winner. Consequently, it ultimately concludes with a tie.

\begin{responsebox}{LLM Response With $T=3.0$}
Assistant A's reasoning contains a critical error---claiming that rolling at least 3 has only a $1/3$ probability, with an incorrect basis that only considers a subset of totals ranging from 2 through 12. This ignores the combinatorially weighted outcomes for each value. The truth (as noted correctly by Assistant B) is that the only impossible case for a sum less than 3 is the sum~$2$, which corresponds to the single combination $(1,1)$ out of a total of 36 possible outcomes. Therefore, $(36 - 1)/36$ gives the final probability as $35/36$.

Furthermore, while Assistant B lists all 35/36 outcomes as acceptable, it initially includes the $(1,1)$ case, thus over‑counting by listing it. However, it clearly notes this incorrect listing and still arrives at the correct result that the probability equals nearly all but one of the 36 outcomes.

Thus both assistants present some flawed thinking. Assistant A, however, makes an incorrect assumption and provides far less precise and misleading data. The true and intended result is that the only impossible total is $2$, so the favorable outcomes correspond to all sums except $2$, giving a probability of $35/36$. Among the two, Assistant~B is preferable for its more structured methodology, thorough enumeration, and effective correction to the correct result.
\end{responsebox}

\subsection{LLM Response With Different Temperature}
\label{sec:llm_response_different_temp}

\subsubsection{Response Example 1}

\begin{responsebox}{LLM Response With $T=0.0$}
Assistant A correctly interprets the reviews and assigns appropriate scores: Review 1 is very
positive (5), but Assistant A gives it a 3, which is incorrect. Review 2 is very negative (1), and
Assistant A gives it a 1, which is correct. Review 3 is neutral/average (3), but Assistant A gives
it a 2, which is slightly low. However, Assistant B's response is completely reversed: Review 1 is
scored as 1 (very negative), which is wrong; Review 2 as 3 (neutral), which is incorrect; and Review
3 as 5 (very positive), which is also wrong. Therefore, Assistant A's answer, while not perfect, is
closer to correct than Assistant B's, which is entirely incorrect.
\end{responsebox}

\begin{responsebox}{LLM Response With $T=0.5$}
Assistant A correctly interprets the reviews and assigns appropriate scores: Review 1 is very
positive (5), but Assistant A assigns 3, which is incorrect. Review 2 is very negative (1),
correctly assigned. Review 3 is neutral (3), but Assistant A assigns 2, which is too low. Assistant
B incorrectly assigns 1 to the first review (which is very positive), 3 to the second (which is very
negative), and 5 to the third (which is neutral). Both assistants make errors, but Assistant A's
scores are closer to the correct interpretation. Review 1 is clearly very positive, so should be 5,
not 3. Review 3 is “okay” and “ordinary,” which is closer to neutral (3) than 2. Therefore,
Assistant A is better.
\end{responsebox}

\begin{responsebox}{LLM Response With $T=1.0$}
Assistant A correctly interprets the sentiment of each review and assigns appropriate scores: Review
1 is very positive (5), but Assistant A assigns 3, which is inaccurate. Review 2 is very negative
(1), and Assistant A assigns 1 correctly. Review 3 is neutral/average (3), but Assistant A assigns
2, which is slightly low. However, Assistant A's scores are closer to the intended sentiment.
Assistant B's response is completely incorrect: they reverse the sentiment values. Review 1 is
highly positive but assigned 1; Review 2 is negative but assigned 3; Review 3 is average but
assigned 5. This indicates a fundamental misunderstanding of the task. Therefore, Assistant A is
better despite minor inaccuracies.
\end{responsebox}

\begin{responsebox}{LLM Response With $T=1.5$}
Assistant A correctly assigned ratings based on the tone and content of each review. Review 1 is
clearly positive, rating 5 is most appropriate, but Assistant A gave it 3, which is neutral—this
is inaccurate. Review 2 is strongly negative, deserving a 1, which Assistant A provided correctly.
Review 3 is lukewarm but slightly positive, so 2 is acceptable. Assistant A's answer is closer to
the correct assessment. Assistant B's answer is completely incorrect: it assigned 1 to review 1
(which is clearly positive), 3 to review 2 (which is clearly negative), and 5 to review 3 (which is
indifferent). This misrepresents all three reviews. Therefore, Assistant A is better, though still
flawed.
\end{responsebox}

\begin{responsebox}{LLM Response With $T=2.0$}
\textit{<missing> Parse Error}
\end{responsebox}

\begin{responsebox}{LLM Human-Like Response With $T=3.0$}
Assistant A correctly assigned ratings based on each review's sentiment. Review 1 (“phenomenal,”
5-point scale) → 5; Review 2 (“unfortunately poor and predictable,” described as “worst”) → 1;
revision of “okay,” “forgettable” → 2, which corresponds to a midrange score (around 2–3 out of 5),
similar to the “middle ground” rating 3. Assistant B incorrectly assigned the scores: Review A,
clearly very positive, deserves 5, while “worst... in 2022” is negative (hence 1). But Assistant B
put Review 2 = 3 (average? neutral?), which was clearly negative. Review 3 was described as mediocre,
so a middle-ground score (2 or 3) would be appropriate. Assistant B, however, put it as 1—the
opposite of what was written.
\end{responsebox}

\subsubsection{Response Example 2}

\begin{responsebox}{LLM Human-Like Response With $T=0.01$}
Assistant B provides a more accurate and nuanced interpretation of Logan's perspective. While
Assistant A correctly identifies revenge as a key theme, the option ``Logan seeks to incite war
against the white settlers'' is too aggressive and misrepresents Logan's complex emotional state.
Logan's speech is primarily about personal grief and the loss of trust in peaceful relations, not a
call to incite war. Assistant B correctly identifies ``Logan laments the loss of his family'' as the
best expression of his perspective, which is supported by the emotional language and key quotes like
``There runs not a drop of my blood in the veins of any living creature.'' This option captures the
core of Logan's lament and emotional state more precisely than the war-inciting interpretation. The
reasoning in B is also more thorough and textually grounded.
\end{responsebox}

\begin{responsebox}{LLM Human-Like Response With $T=0.5$}
Assistant B provides a more accurate and nuanced interpretation of Logan's perspective. While
Assistant A correctly identifies that Logan seeks revenge, the option ``Logan seeks to incite war
against the white settlers'' is too aggressive and misrepresents the tone of the passage. Logan does
not advocate for ongoing war; rather, he expresses personal vengeance and a desire for peace after
satisfying his revenge. Assistant B correctly identifies ``Logan laments the loss of his family'' as
the best expression of his perspective, which captures the emotional core of the passage---grief,
loss, and the tragic end of his trust in the white settlers. The reasoning in B is also more
thorough and contextually grounded in the text. Therefore, B is superior in accuracy, depth, and
relevance.
\end{responsebox}

\begin{responsebox}{LLM Human-Like Response With $T=1.0$}
Assistant B provides a more accurate and nuanced interpretation of Logan's perspective. While
Assistant A correctly identifies that Logan seeks revenge, its choice of ``Logan seeks to incite war
against the white settlers'' misrepresents the tone and intent of the passage. Logan's statement is a
lament of personal loss and grief, not an incitement to war. The passage shows Logan has already
carried out revenge and now expresses relief at the prospect of peace, stating ``For my country, I
rejoice at the beams of peace.'' Assistant B correctly identifies ``Logan laments the loss of his
family'' as the central emotional core, which is supported by direct evidence from the text.
Additionally, Assistant B's response is clearer, more coherent, and better aligned with the text's
emotional arc.
\end{responsebox}

\begin{responsebox}{LLM Human-Like Response With $T=1.5$}
Assistant B provides a more accurate and nuanced interpretation of Logan's perspective. While
Assistant A correctly identifies that Logan seeks revenge, its choice of ``Logan seeks to incite war
against the white settlers'' misrepresents the tone and focus of the passage. Logan explicitly states
he is not driven by fear and that he rejoices at the beams of peace, indicating he does not want to
continue or incite war. His revenge was personal and complete, not a call to ongoing conflict.
Assistant B correctly identifies the central emotional core of the passage: Logan's lament for his
lost family, which is the foundational motivation for his actions and the main theme of the text.
This is supported by multiple key phrases in the passage, such as the line about ``no drop of my
blood in the veins of any living creature,'' which expresses profound grief. Assistant A's choice
reflects a misunderstanding of Logan's ultimate stance on peace, while Assistant B accurately
captures the deep personal sorrow that drives Logan's perspective.
\end{responsebox}

\begin{responsebox}{LLM Human-Like Response With $T=2.0$}
Assistant B provides a more accurate and nuanced interpretation of Logan's perspective. While
Assistant A correctly identifies that Logan seeks revenge, its chosen option ``Logan seeks to incite
war against the white settlers'' misrepresents the tone and focus of the passage. Logan explicitly
states he is not driven by fear and that he does not wish to live in hatred, ultimately finding joy
in the ``beams of peace.'' The passage shows that his revenge is a response to personal tragedy, not
an aggressive political objective. The central emotional core is his lamentation over lost family,
captured clearly in Assistant B's answer. Assistant B also correctly references relevant textual
evidence (``There runs not a drop of my blood in the veins\ldots''), supporting the conclusion with
greater sensitivity to thematic depth and emotional authenticity. Therefore, Assistant B is better.
\end{responsebox}

\begin{responsebox}{LLM Human-Like Response With $T=3.0$}
Assistant B provides a more accurate, nuanced, and thorough justification for their choice,
identifying the primary thematic core (deep grief and personal tragedy). This is a foundational
component emphasized through emotional statements and poetic phrasing, particularly after
establishing the context of past alliances and peaceful gestures.
The key detail---his loved ones' annihilation being complete enough to leave his blood extinct in
existence---is accurately cited.
Assistant A's response incorrectly selects ``Logan seeks to incite war.'' While Logan does pursue
revenge, the central emotional purpose of his address lies more deeply embedded around mourning
than incitement. There is no call-to-arms, no urging of others to act, which the selection falsely
represents. Rather, the vindicating justice narrative centers around the tragedy he experienced
directly, making it the focus and hence the best choice for capturing the primary emotional thrust.
Assistant A's reasoning is fundamentally flawed: the chosen option misrepresents Logan's ultimate
stance, while Assistant B accurately captures the deep personal sorrow driving Logan's perspective.
\end{responsebox}

\section{Vllm Inference Engine}

To achieve the high throughput required for our experiments, we implemented an optimized asynchronous inference pipeline. The architecture employs a hierarchical concurrency model: a main loop orchestrates task scheduling, controlled by an \texttt{asyncio.Semaphore} limiting concurrent batches to \texttt{MAX\_CONCURRENT\_BATCHES=8}. Each batch (\texttt{BATCH\_SIZE=64}) is processed via \texttt{semaphore}, leveraging vLLM's native batch inference through a single API call to \texttt{batch\_generate\_prompts/completions}. This design delivers exceptional performance: $64 \times 8 = 512$ concurrent tokens---over 64$\times$ faster than synchronous processing---enabled by vLLM's KV cache optimization and parallel decoding which substantially accelerating our experimental pipeline. Each experimental condition explicitly varies the random seed to ensure reproducibility. The overall computation budget is estimated at approximately 1024 H100 GPU-hours.

For \textbf{temperature annealing method}, we implement token-level temperature annealing using a custom Logits Processor in the vLLM inference engine. This plugin interface allows the logits of the entire batch to be modified directly before each sampling step. We implement a Token-Level Temperature Annealing Logits Processor that runs before vLLM's built-in temperature scaling.

\section{License Statement}

All models employed in this study are publicly available under open or community licenses. Specifically, Qwen3-30B and Qwen3-Next-80B are released under the Apache License 2.0, which permits free use, modification, and redistribution for both research and commercial purposes. Llama-3.1-8B-Instruct is governed by the Meta Llama 3.1 Community License Agreement, allowing research and commercial use subject to specific constraints. gemma-3-27B is made available under the Google Gemma Terms of Use, requiring users to agree to Google's usage policy prior to access. Regarding evaluation benchmarks, MT-Bench Human Judgments is released under the Creative Commons Attribution 4.0 International License (CC BY 4.0), while MMLU-Pro is distributed under the MIT License, both permitting academic use with appropriate attribution.

\section{Additional Results}

Tables~\ref{tab:appendix_full_results_1}, \ref{tab:appendix_full_results_2}, and \ref{tab:appendix_full_results_3} report results across all temperatures and additionally include Qwen3.8-Flash-Next-FP8 for reference; this model supports temperatures up to $2.0$ only.

\definecolor{TColdBg}{RGB}{233,245,255} 
\definecolor{TCoolBg}{RGB}{233,247,247} 
\definecolor{TMidBg}{RGB}{236,247,239}  
\definecolor{TMildBg}{RGB}{245,248,230} 
\definecolor{TWarmBg}{RGB}{255,247,226} 
\definecolor{THotBg}{RGB}{255,242,232}  
\definecolor{TColdText}{RGB}{0,92,171}
\definecolor{TCoolText}{RGB}{0,118,128}
\definecolor{TMidText}{RGB}{20,115,60}
\definecolor{TMildText}{RGB}{105,125,10}
\definecolor{TWarmText}{RGB}{190,115,0}
\definecolor{THotText}{RGB}{176,60,0}

\begin{table*}[htbp]
\centering
\begin{adjustbox}{width=1.00\textwidth}
\begin{tabular}{c>{\raggedright\arraybackslash}p{3.25cm}cccccccc}
\toprule
\multirow{2}{*}{\textbf{\faBalanceScale T \faBalanceScale}} & \multirow{2}{*}{\textbf{Model}} & \multirow{2}{*}{\textbf{Judge Type}} & \multirow{2}{*}{\textbf{\begin{tabular}[c]{@{}c@{}}Prompt\\ Strategy\end{tabular}}} & \multicolumn{3}{c}{\textbf{MT-Bench Human Judgments}} & \multicolumn{3}{c}{\textbf{MMLU\_Pro}} \\
\cmidrule(lr){5-7}\cmidrule(lr){8-10}
& & & & \textbf{\begin{tabular}[c]{@{}c@{}}AGR$\uparrow$ \\ ($\mu \pm$ std)\end{tabular}} & \textbf{\begin{tabular}[c]{@{}c@{}}CON$\uparrow$ \\ ($\mu \pm$ std)\end{tabular}} & \textbf{\begin{tabular}[c]{@{}c@{}}ERR$\downarrow$ \\ ($\mu \pm$ std)\end{tabular}} & \textbf{\begin{tabular}[c]{@{}c@{}}AGR$\uparrow$ \\ ($\mu \pm$ std)\end{tabular}} & \textbf{\begin{tabular}[c]{@{}c@{}}CON$\uparrow$ \\ ($\mu \pm$ std)\end{tabular}} & \textbf{\begin{tabular}[c]{@{}c@{}}ERR$\downarrow$ \\ ($\mu \pm$ std)\end{tabular}} \\
\midrule

\cellcolor{TColdBg} & \cellcolor{TColdBg} & \cellcolor{TColdBg} & \cellcolor{TColdBg}Direct & \cellcolor{TColdBg}0.61 $\pm$ 0.00 & \cellcolor{TColdBg}0.98 $\pm$ 0.09 & \cellcolor{TColdBg}0.00 $\pm$ 0.00 & \cellcolor{TColdBg}0.58 $\pm$ 0.01 & \cellcolor{TColdBg}0.97 $\pm$ 0.12 & \cellcolor{TColdBg}0.00 $\pm$ 0.00 \\
\cellcolor{TColdBg} & \cellcolor{TColdBg} & \cellcolor{TColdBg}\multirow{-2}{*}{\begin{tabular}[c]{@{}c@{}}Pairwise\\Comparison\end{tabular}} & \cellcolor{TColdBg}CoT & \cellcolor{TColdBg}0.59 $\pm$ 0.01 & \cellcolor{TColdBg}0.94 $\pm$ 0.16 & \cellcolor{TColdBg}0.04 $\pm$ 0.01 & \cellcolor{TColdBg}0.55 $\pm$ 0.02 & \cellcolor{TColdBg}0.95 $\pm$ 0.16 & \cellcolor{TColdBg}0.25 $\pm$ 0.03 \\
\cellcolor{TColdBg} & \cellcolor{TColdBg} & \cellcolor{TColdBg} & \cellcolor{TColdBg}Direct & \cellcolor{TColdBg}0.54 $\pm$ 0.01 & \cellcolor{TColdBg}0.95 $\pm$ 0.14 & \cellcolor{TColdBg}0.00 $\pm$ 0.00 & \cellcolor{TColdBg}0.70 $\pm$ 0.01 & \cellcolor{TColdBg}0.94 $\pm$ 0.16 & \cellcolor{TColdBg}0.00 $\pm$ 0.00 \\
\cellcolor{TColdBg} & \cellcolor{TColdBg} & \cellcolor{TColdBg}\multirow{-2}{*}{\begin{tabular}[c]{@{}c@{}}Single Answer\\Grading\end{tabular}} & \cellcolor{TColdBg}CoT & \cellcolor{TColdBg}0.54 $\pm$ 0.01 & \cellcolor{TColdBg}0.91 $\pm$ 0.18 & \cellcolor{TColdBg}0.01 $\pm$ 0.00 & \cellcolor{TColdBg}0.63 $\pm$ 0.02 & \cellcolor{TColdBg}0.83 $\pm$ 0.26 & \cellcolor{TColdBg}0.17 $\pm$ 0.02 \\
\cellcolor{TColdBg} & \cellcolor{TColdBg} & \cellcolor{TColdBg} & \cellcolor{TColdBg}Direct & \cellcolor{TColdBg}0.58 $\pm$ 0.01 & \cellcolor{TColdBg}0.94 $\pm$ 0.15 & \cellcolor{TColdBg}0.00 $\pm$ 0.00 & \cellcolor{TColdBg}0.65 $\pm$ 0.01 & \cellcolor{TColdBg}0.94 $\pm$ 0.16 & \cellcolor{TColdBg}0.00 $\pm$ 0.00 \\
\cellcolor{TColdBg} & \cellcolor{TColdBg}\multirow{-6}{*}{\begin{tabular}[c]{@{}l@{}}Qwen/Qwen3-Next\\80B-A3B-Instruct\\FP8\end{tabular}} & \cellcolor{TColdBg}\multirow{-2}{*}{\begin{tabular}[c]{@{}c@{}}Reference Guided\\Grading\end{tabular}} & \cellcolor{TColdBg}CoT & \cellcolor{TColdBg}0.58 $\pm$ 0.01 & \cellcolor{TColdBg}0.90 $\pm$ 0.20 & \cellcolor{TColdBg}0.05 $\pm$ 0.01 & \cellcolor{TColdBg}0.64 $\pm$ 0.02 & \cellcolor{TColdBg}0.86 $\pm$ 0.22 & \cellcolor{TColdBg}0.11 $\pm$ 0.01 \\
\cellcolor{TColdBg} & \cellcolor{TColdBg} & \cellcolor{TColdBg} & \cellcolor{TColdBg}Direct & \cellcolor{TColdBg}0.59 $\pm$ 0.00 & \cellcolor{TColdBg}\textbf{1.00 $\pm$ 0.04} & \cellcolor{TColdBg}0.00 $\pm$ 0.00 & \cellcolor{TColdBg}0.61 $\pm$ 0.00 & \cellcolor{TColdBg}0.99 $\pm$ 0.08 & \cellcolor{TColdBg}0.00 $\pm$ 0.00 \\
\cellcolor{TColdBg} & \cellcolor{TColdBg} & \cellcolor{TColdBg}\multirow{-2}{*}{\begin{tabular}[c]{@{}c@{}}Pairwise\\Comparison\end{tabular}} & \cellcolor{TColdBg}CoT & \cellcolor{TColdBg}0.56 $\pm$ 0.00 & \cellcolor{TColdBg}0.99 $\pm$ 0.08 & \cellcolor{TColdBg}0.00 $\pm$ 0.00 & \cellcolor{TColdBg}0.53 $\pm$ 0.01 & \cellcolor{TColdBg}0.98 $\pm$ 0.11 & \cellcolor{TColdBg}0.00 $\pm$ 0.00 \\
\cellcolor{TColdBg} & \cellcolor{TColdBg} & \cellcolor{TColdBg} & \cellcolor{TColdBg}Direct & \cellcolor{TColdBg}0.51 $\pm$ 0.00 & \cellcolor{TColdBg}0.98 $\pm$ 0.09 & \cellcolor{TColdBg}0.00 $\pm$ 0.00 & \cellcolor{TColdBg}0.56 $\pm$ 0.01 & \cellcolor{TColdBg}0.98 $\pm$ 0.11 & \cellcolor{TColdBg}0.00 $\pm$ 0.00 \\
\cellcolor{TColdBg} & \cellcolor{TColdBg} & \cellcolor{TColdBg}\multirow{-2}{*}{\begin{tabular}[c]{@{}c@{}}Single Answer\\Grading\end{tabular}} & \cellcolor{TColdBg}CoT & \cellcolor{TColdBg}0.48 $\pm$ 0.01 & \cellcolor{TColdBg}0.94 $\pm$ 0.15 & \cellcolor{TColdBg}0.00 $\pm$ 0.00 & \cellcolor{TColdBg}0.64 $\pm$ 0.01 & \cellcolor{TColdBg}0.91 $\pm$ 0.18 & \cellcolor{TColdBg}0.00 $\pm$ 0.00 \\
\cellcolor{TColdBg} & \cellcolor{TColdBg} & \cellcolor{TColdBg} & \cellcolor{TColdBg}Direct & \cellcolor{TColdBg}0.57 $\pm$ 0.00 & \cellcolor{TColdBg}\textbf{1.00 $\pm$ 0.04} & \cellcolor{TColdBg}0.00 $\pm$ 0.00 & \cellcolor{TColdBg}0.69 $\pm$ 0.01 & \cellcolor{TColdBg}0.99 $\pm$ 0.09 & \cellcolor{TColdBg}0.00 $\pm$ 0.00 \\
\cellcolor{TColdBg} & \cellcolor{TColdBg}\multirow{-6}{*}{\begin{tabular}[c]{@{}l@{}}google/\\gemma-3-27b-it\end{tabular}} & \cellcolor{TColdBg}\multirow{-2}{*}{\begin{tabular}[c]{@{}c@{}}Reference Guided\\Grading\end{tabular}} & \cellcolor{TColdBg}CoT & \cellcolor{TColdBg}0.57 $\pm$ 0.01 & \cellcolor{TColdBg}0.96 $\pm$ 0.13 & \cellcolor{TColdBg}0.00 $\pm$ 0.00 & \cellcolor{TColdBg}0.68 $\pm$ 0.01 & \cellcolor{TColdBg}0.95 $\pm$ 0.15 & \cellcolor{TColdBg}0.00 $\pm$ 0.00 \\
\cellcolor{TColdBg} & \cellcolor{TColdBg} & \cellcolor{TColdBg} & \cellcolor{TColdBg}Direct & \cellcolor{TColdBg}0.65 $\pm$ 0.05 & \cellcolor{TColdBg}0.99 $\pm$ 0.06 & \cellcolor{TColdBg}0.90 $\pm$ 0.02 & \cellcolor{TColdBg}0.75 $\pm$ 0.06 & \cellcolor{TColdBg}0.92 $\pm$ 0.25 & \cellcolor{TColdBg}0.89 $\pm$ 0.02 \\
\cellcolor{TColdBg} & \cellcolor{TColdBg} & \cellcolor{TColdBg}\multirow{-2}{*}{\begin{tabular}[c]{@{}c@{}}Pairwise\\Comparison\end{tabular}} & \cellcolor{TColdBg}CoT & \cellcolor{TColdBg}0.66 $\pm$ 0.05 & \cellcolor{TColdBg}0.94 $\pm$ 0.20 & \cellcolor{TColdBg}0.77 $\pm$ 0.02 & \cellcolor{TColdBg}0.69 $\pm$ 0.07 & \cellcolor{TColdBg}0.91 $\pm$ 0.21 & \cellcolor{TColdBg}0.79 $\pm$ 0.03 \\
\cellcolor{TColdBg} & \cellcolor{TColdBg} & \cellcolor{TColdBg} & \cellcolor{TColdBg}Direct & \cellcolor{TColdBg}0.42 $\pm$ 0.29 & \cellcolor{TColdBg}\textbf{1.00 $\pm$ 0.00} & \cellcolor{TColdBg}0.99 $\pm$ 0.00 & \cellcolor{TColdBg}0.50 $\pm$ 0.71 & \cellcolor{TColdBg}nan $\pm$ 0.00 & \cellcolor{TColdBg}1.00 $\pm$ 0.00 \\
\cellcolor{TColdBg} & \cellcolor{TColdBg} & \cellcolor{TColdBg}\multirow{-2}{*}{\begin{tabular}[c]{@{}c@{}}Single Answer\\Grading\end{tabular}} & \cellcolor{TColdBg}CoT & \cellcolor{TColdBg}0.61 $\pm$ 0.05 & \cellcolor{TColdBg}0.81 $\pm$ 0.32 & \cellcolor{TColdBg}0.86 $\pm$ 0.01 & \cellcolor{TColdBg}0.75 $\pm$ 0.07 & \cellcolor{TColdBg}0.85 $\pm$ 0.29 & \cellcolor{TColdBg}0.89 $\pm$ 0.01 \\
\cellcolor{TColdBg} & \cellcolor{TColdBg} & \cellcolor{TColdBg} & \cellcolor{TColdBg}Direct & \cellcolor{TColdBg}\textbf{1.00 $\pm$ 0.00} & \cellcolor{TColdBg}\textbf{1.00 $\pm$ 0.00} & \cellcolor{TColdBg}0.99 $\pm$ 0.00 & \cellcolor{TColdBg}\textbf{1.00 $\pm$ 0.00} & \cellcolor{TColdBg}\textbf{1.00 $\pm$ 0.00} & \cellcolor{TColdBg}0.99 $\pm$ 0.00 \\
\cellcolor{TColdBg}\multirow{-18}{*}{\textbf{\textcolor{TColdText}{\faSnowflake} 0.01 \textcolor{TColdText}{\faSnowflake}}} & \cellcolor{TColdBg}\multirow{-6}{*}{\begin{tabular}[c]{@{}l@{}}Qwen3.8-Flash-Next\\FP8\end{tabular}} & \cellcolor{TColdBg}\multirow{-2}{*}{\begin{tabular}[c]{@{}c@{}}Reference Guided\\Grading\end{tabular}} & \cellcolor{TColdBg}CoT & \cellcolor{TColdBg}0.59 $\pm$ 0.08 & \cellcolor{TColdBg}0.90 $\pm$ 0.25 & \cellcolor{TColdBg}0.95 $\pm$ 0.01 & \cellcolor{TColdBg}0.65 $\pm$ 0.11 & \cellcolor{TColdBg}0.90 $\pm$ 0.19 & \cellcolor{TColdBg}0.94 $\pm$ 0.01 \\

\midrule

\cellcolor{TCoolBg} & \cellcolor{TCoolBg} & \cellcolor{TCoolBg} & \cellcolor{TCoolBg}Direct & \cellcolor{TCoolBg}0.61 $\pm$ 0.01 & \cellcolor{TCoolBg}0.97 $\pm$ 0.12 & \cellcolor{TCoolBg}0.00 $\pm$ 0.00 & \cellcolor{TCoolBg}0.57 $\pm$ 0.01 & \cellcolor{TCoolBg}0.96 $\pm$ 0.15 & \cellcolor{TCoolBg}0.00 $\pm$ 0.00 \\
\cellcolor{TCoolBg} & \cellcolor{TCoolBg} & \cellcolor{TCoolBg}\multirow{-2}{*}{\begin{tabular}[c]{@{}c@{}}Pairwise\\Comparison\end{tabular}} & \cellcolor{TCoolBg}CoT & \cellcolor{TCoolBg}0.59 $\pm$ 0.01 & \cellcolor{TCoolBg}0.93 $\pm$ 0.17 & \cellcolor{TCoolBg}0.04 $\pm$ 0.03 & \cellcolor{TCoolBg}0.54 $\pm$ 0.03 & \cellcolor{TCoolBg}0.92 $\pm$ 0.18 & \cellcolor{TCoolBg}0.26 $\pm$ 0.06 \\
\cellcolor{TCoolBg} & \cellcolor{TCoolBg} & \cellcolor{TCoolBg} & \cellcolor{TCoolBg}Direct & \cellcolor{TCoolBg}0.54 $\pm$ 0.01 & \cellcolor{TCoolBg}0.92 $\pm$ 0.18 & \cellcolor{TCoolBg}0.00 $\pm$ 0.00 & \cellcolor{TCoolBg}0.66 $\pm$ 0.02 & \cellcolor{TCoolBg}0.85 $\pm$ 0.22 & \cellcolor{TCoolBg}0.00 $\pm$ 0.00 \\
\cellcolor{TCoolBg} & \cellcolor{TCoolBg} & \cellcolor{TCoolBg}\multirow{-2}{*}{\begin{tabular}[c]{@{}c@{}}Single Answer\\Grading\end{tabular}} & \cellcolor{TCoolBg}CoT & \cellcolor{TCoolBg}0.54 $\pm$ 0.01 & \cellcolor{TCoolBg}0.87 $\pm$ 0.22 & \cellcolor{TCoolBg}0.02 $\pm$ 0.01 & \cellcolor{TCoolBg}0.62 $\pm$ 0.04 & \cellcolor{TCoolBg}0.78 $\pm$ 0.27 & \cellcolor{TCoolBg}0.17 $\pm$ 0.03 \\
\cellcolor{TCoolBg} & \cellcolor{TCoolBg} & \cellcolor{TCoolBg} & \cellcolor{TCoolBg}Direct & \cellcolor{TCoolBg}0.57 $\pm$ 0.01 & \cellcolor{TCoolBg}0.92 $\pm$ 0.18 & \cellcolor{TCoolBg}0.00 $\pm$ 0.00 & \cellcolor{TCoolBg}0.64 $\pm$ 0.01 & \cellcolor{TCoolBg}0.90 $\pm$ 0.21 & \cellcolor{TCoolBg}0.00 $\pm$ 0.00 \\
\cellcolor{TCoolBg} & \cellcolor{TCoolBg}\multirow{-6}{*}{\begin{tabular}[c]{@{}l@{}}Qwen/Qwen3-Next\\80B-A3B-Instruct\\FP8\end{tabular}} & \cellcolor{TCoolBg}\multirow{-2}{*}{\begin{tabular}[c]{@{}c@{}}Reference Guided\\Grading\end{tabular}} & \cellcolor{TCoolBg}CoT & \cellcolor{TCoolBg}0.57 $\pm$ 0.01 & \cellcolor{TCoolBg}0.88 $\pm$ 0.21 & \cellcolor{TCoolBg}0.06 $\pm$ 0.03 & \cellcolor{TCoolBg}0.62 $\pm$ 0.03 & \cellcolor{TCoolBg}0.85 $\pm$ 0.23 & \cellcolor{TCoolBg}0.13 $\pm$ 0.05 \\
\cellcolor{TCoolBg} & \cellcolor{TCoolBg} & \cellcolor{TCoolBg} & \cellcolor{TCoolBg}Direct & \cellcolor{TCoolBg}0.59 $\pm$ 0.00 & \cellcolor{TCoolBg}0.99 $\pm$ 0.07 & \cellcolor{TCoolBg}0.00 $\pm$ 0.00 & \cellcolor{TCoolBg}0.61 $\pm$ 0.00 & \cellcolor{TCoolBg}0.99 $\pm$ 0.10 & \cellcolor{TCoolBg}0.00 $\pm$ 0.00 \\
\cellcolor{TCoolBg} & \cellcolor{TCoolBg} & \cellcolor{TCoolBg}\multirow{-2}{*}{\begin{tabular}[c]{@{}c@{}}Pairwise\\Comparison\end{tabular}} & \cellcolor{TCoolBg}CoT & \cellcolor{TCoolBg}0.56 $\pm$ 0.00 & \cellcolor{TCoolBg}0.97 $\pm$ 0.11 & \cellcolor{TCoolBg}0.00 $\pm$ 0.00 & \cellcolor{TCoolBg}0.54 $\pm$ 0.01 & \cellcolor{TCoolBg}0.96 $\pm$ 0.14 & \cellcolor{TCoolBg}0.00 $\pm$ 0.00 \\
\cellcolor{TCoolBg} & \cellcolor{TCoolBg} & \cellcolor{TCoolBg} & \cellcolor{TCoolBg}Direct & \cellcolor{TCoolBg}0.51 $\pm$ 0.01 & \cellcolor{TCoolBg}0.96 $\pm$ 0.14 & \cellcolor{TCoolBg}0.00 $\pm$ 0.00 & \cellcolor{TCoolBg}0.56 $\pm$ 0.01 & \cellcolor{TCoolBg}0.93 $\pm$ 0.19 & \cellcolor{TCoolBg}0.00 $\pm$ 0.00 \\
\cellcolor{TCoolBg} & \cellcolor{TCoolBg} & \cellcolor{TCoolBg}\multirow{-2}{*}{\begin{tabular}[c]{@{}c@{}}Single Answer\\Grading\end{tabular}} & \cellcolor{TCoolBg}CoT & \cellcolor{TCoolBg}0.47 $\pm$ 0.01 & \cellcolor{TCoolBg}0.85 $\pm$ 0.22 & \cellcolor{TCoolBg}0.00 $\pm$ 0.00 & \cellcolor{TCoolBg}0.62 $\pm$ 0.02 & \cellcolor{TCoolBg}0.84 $\pm$ 0.23 & \cellcolor{TCoolBg}0.00 $\pm$ 0.00 \\
\cellcolor{TCoolBg} & \cellcolor{TCoolBg} & \cellcolor{TCoolBg} & \cellcolor{TCoolBg}Direct & \cellcolor{TCoolBg}0.57 $\pm$ 0.00 & \cellcolor{TCoolBg}0.99 $\pm$ 0.09 & \cellcolor{TCoolBg}0.00 $\pm$ 0.00 & \cellcolor{TCoolBg}0.68 $\pm$ 0.01 & \cellcolor{TCoolBg}0.97 $\pm$ 0.14 & \cellcolor{TCoolBg}0.00 $\pm$ 0.00 \\
\cellcolor{TCoolBg} & \cellcolor{TCoolBg}\multirow{-6}{*}{\begin{tabular}[c]{@{}l@{}}google/\\gemma-3-27b-it\end{tabular}} & \cellcolor{TCoolBg}\multirow{-2}{*}{\begin{tabular}[c]{@{}c@{}}Reference Guided\\Grading\end{tabular}} & \cellcolor{TCoolBg}CoT & \cellcolor{TCoolBg}0.56 $\pm$ 0.01 & \cellcolor{TCoolBg}0.92 $\pm$ 0.19 & \cellcolor{TCoolBg}0.00 $\pm$ 0.00 & \cellcolor{TCoolBg}0.68 $\pm$ 0.02 & \cellcolor{TCoolBg}0.90 $\pm$ 0.18 & \cellcolor{TCoolBg}0.00 $\pm$ 0.00 \\
\cellcolor{TCoolBg} & \cellcolor{TCoolBg} & \cellcolor{TCoolBg} & \cellcolor{TCoolBg}Direct & \cellcolor{TCoolBg}0.66 $\pm$ 0.07 & \cellcolor{TCoolBg}0.89 $\pm$ 0.29 & \cellcolor{TCoolBg}0.88 $\pm$ 0.01 & \cellcolor{TCoolBg}0.68 $\pm$ 0.09 & \cellcolor{TCoolBg}0.89 $\pm$ 0.24 & \cellcolor{TCoolBg}0.86 $\pm$ 0.03 \\
\cellcolor{TCoolBg} & \cellcolor{TCoolBg} & \cellcolor{TCoolBg}\multirow{-2}{*}{\begin{tabular}[c]{@{}c@{}}Pairwise\\Comparison\end{tabular}} & \cellcolor{TCoolBg}CoT & \cellcolor{TCoolBg}0.66 $\pm$ 0.05 & \cellcolor{TCoolBg}0.89 $\pm$ 0.26 & \cellcolor{TCoolBg}0.74 $\pm$ 0.02 & \cellcolor{TCoolBg}0.71 $\pm$ 0.06 & \cellcolor{TCoolBg}0.86 $\pm$ 0.28 & \cellcolor{TCoolBg}0.77 $\pm$ 0.04 \\
\cellcolor{TCoolBg} & \cellcolor{TCoolBg} & \cellcolor{TCoolBg} & \cellcolor{TCoolBg}Direct & \cellcolor{TCoolBg}0.48 $\pm$ 0.36 & \cellcolor{TCoolBg}\textbf{1.00 $\pm$ 0.00} & \cellcolor{TCoolBg}0.99 $\pm$ 0.00 & \cellcolor{TCoolBg}0.10 $\pm$ 0.22 & \cellcolor{TCoolBg}\textbf{1.00 $\pm$ 0.00} & \cellcolor{TCoolBg}1.00 $\pm$ 0.00 \\
\cellcolor{TCoolBg} & \cellcolor{TCoolBg} & \cellcolor{TCoolBg}\multirow{-2}{*}{\begin{tabular}[c]{@{}c@{}}Single Answer\\Grading\end{tabular}} & \cellcolor{TCoolBg}CoT & \cellcolor{TCoolBg}0.60 $\pm$ 0.06 & \cellcolor{TCoolBg}0.80 $\pm$ 0.33 & \cellcolor{TCoolBg}0.85 $\pm$ 0.02 & \cellcolor{TCoolBg}0.71 $\pm$ 0.10 & \cellcolor{TCoolBg}0.65 $\pm$ 0.40 & \cellcolor{TCoolBg}0.87 $\pm$ 0.02 \\
\cellcolor{TCoolBg} & \cellcolor{TCoolBg} & \cellcolor{TCoolBg} & \cellcolor{TCoolBg}Direct & \cellcolor{TCoolBg}\textbf{0.70 $\pm$ 0.14} & \cellcolor{TCoolBg}0.93 $\pm$ 0.26 & \cellcolor{TCoolBg}0.98 $\pm$ 0.01 & \cellcolor{TCoolBg}\textbf{0.76 $\pm$ 0.22} & \cellcolor{TCoolBg}0.97 $\pm$ 0.11 & \cellcolor{TCoolBg}0.98 $\pm$ 0.01 \\
\cellcolor{TCoolBg}\multirow{-18}{*}{\textbf{\textcolor{TCoolText}{\faThermometerQuarter} 0.5 \textcolor{TCoolText}{\faThermometerQuarter}}} & \cellcolor{TCoolBg}\multirow{-6}{*}{\begin{tabular}[c]{@{}l@{}}Qwen3.8-Flash-Next\\FP8\end{tabular}} & \cellcolor{TCoolBg}\multirow{-2}{*}{\begin{tabular}[c]{@{}c@{}}Reference Guided\\Grading\end{tabular}} & \cellcolor{TCoolBg}CoT & \cellcolor{TCoolBg}0.63 $\pm$ 0.08 & \cellcolor{TCoolBg}0.86 $\pm$ 0.29 & \cellcolor{TCoolBg}0.92 $\pm$ 0.01 & \cellcolor{TCoolBg}0.73 $\pm$ 0.09 & \cellcolor{TCoolBg}0.80 $\pm$ 0.37 & \cellcolor{TCoolBg}0.92 $\pm$ 0.01 \\
\bottomrule
\end{tabular}
\end{adjustbox}
\caption{Full results of LLM-as-a-Judge stability for $T \in \{0.01, 0.5\}$ (part 1 of 3; all temperatures $T \in \{0.01, 0.5, 1.0, 1.5, 2.0, 3.0\}$ are reported across Tables~\ref{tab:appendix_full_results_1}--\ref{tab:appendix_full_results_3}). Three judge models, three evaluation paradigms (Pairwise Comparison, Single Answer Grading, and Reference-Guided Grading), and two prompting strategies (Direct and CoT), measured by agreement with human judgments (AGR; $\mu \pm$ std over 10 repeats), 1-flip consistency across repeats (CON; $\mu \pm$ std over questions), and format error rate (ERR; $\mu \pm$ std over 10 repeats) on MT-Bench and MMLU-Pro. \textbf{Bold} marks the best AGR/CON (highest) and the uniquely lowest ERR within each temperature block and dataset.}
\label{tab:appendix_full_results_1}
\end{table*}

\begin{table*}[htbp]
\centering
\begin{adjustbox}{width=1.00\textwidth}
\begin{tabular}{c>{\raggedright\arraybackslash}p{3.25cm}cccccccc}
\toprule
\multirow{2}{*}{\textbf{\faBalanceScale T \faBalanceScale}} & \multirow{2}{*}{\textbf{Model}} & \multirow{2}{*}{\textbf{Judge Type}} & \multirow{2}{*}{\textbf{\begin{tabular}[c]{@{}c@{}}Prompt\\ Strategy\end{tabular}}} & \multicolumn{3}{c}{\textbf{MT-Bench Human Judgments}} & \multicolumn{3}{c}{\textbf{MMLU\_Pro}} \\
\cmidrule(lr){5-7}\cmidrule(lr){8-10}
& & & & \textbf{\begin{tabular}[c]{@{}c@{}}AGR$\uparrow$ \\ ($\mu \pm$ std)\end{tabular}} & \textbf{\begin{tabular}[c]{@{}c@{}}CON$\uparrow$ \\ ($\mu \pm$ std)\end{tabular}} & \textbf{\begin{tabular}[c]{@{}c@{}}ERR$\downarrow$ \\ ($\mu \pm$ std)\end{tabular}} & \textbf{\begin{tabular}[c]{@{}c@{}}AGR$\uparrow$ \\ ($\mu \pm$ std)\end{tabular}} & \textbf{\begin{tabular}[c]{@{}c@{}}CON$\uparrow$ \\ ($\mu \pm$ std)\end{tabular}} & \textbf{\begin{tabular}[c]{@{}c@{}}ERR$\downarrow$ \\ ($\mu \pm$ std)\end{tabular}} \\
\midrule

\cellcolor{TMidBg} & \cellcolor{TMidBg} & \cellcolor{TMidBg} & \cellcolor{TMidBg}Direct & \cellcolor{TMidBg}0.62 $\pm$ 0.01 & \cellcolor{TMidBg}0.96 $\pm$ 0.13 & \cellcolor{TMidBg}0.00 $\pm$ 0.00 & \cellcolor{TMidBg}0.58 $\pm$ 0.02 & \cellcolor{TMidBg}0.93 $\pm$ 0.19 & \cellcolor{TMidBg}0.00 $\pm$ 0.00 \\
\cellcolor{TMidBg} & \cellcolor{TMidBg} & \cellcolor{TMidBg}\multirow{-2}{*}{\begin{tabular}[c]{@{}c@{}}Pairwise\\Comparison\end{tabular}} & \cellcolor{TMidBg}CoT & \cellcolor{TMidBg}0.59 $\pm$ 0.01 & \cellcolor{TMidBg}0.90 $\pm$ 0.21 & \cellcolor{TMidBg}0.06 $\pm$ 0.05 & \cellcolor{TMidBg}0.54 $\pm$ 0.04 & \cellcolor{TMidBg}0.87 $\pm$ 0.25 & \cellcolor{TMidBg}0.28 $\pm$ 0.12 \\
\cellcolor{TMidBg} & \cellcolor{TMidBg} & \cellcolor{TMidBg} & \cellcolor{TMidBg}Direct & \cellcolor{TMidBg}0.54 $\pm$ 0.01 & \cellcolor{TMidBg}0.87 $\pm$ 0.21 & \cellcolor{TMidBg}0.00 $\pm$ 0.00 & \cellcolor{TMidBg}0.64 $\pm$ 0.03 & \cellcolor{TMidBg}0.82 $\pm$ 0.24 & \cellcolor{TMidBg}0.00 $\pm$ 0.00 \\
\cellcolor{TMidBg} & \cellcolor{TMidBg} & \cellcolor{TMidBg}\multirow{-2}{*}{\begin{tabular}[c]{@{}c@{}}Single Answer\\Grading\end{tabular}} & \cellcolor{TMidBg}CoT & \cellcolor{TMidBg}0.53 $\pm$ 0.01 & \cellcolor{TMidBg}0.82 $\pm$ 0.24 & \cellcolor{TMidBg}0.02 $\pm$ 0.01 & \cellcolor{TMidBg}0.62 $\pm$ 0.05 & \cellcolor{TMidBg}0.73 $\pm$ 0.30 & \cellcolor{TMidBg}0.18 $\pm$ 0.08 \\
\cellcolor{TMidBg} & \cellcolor{TMidBg} & \cellcolor{TMidBg} & \cellcolor{TMidBg}Direct & \cellcolor{TMidBg}0.56 $\pm$ 0.02 & \cellcolor{TMidBg}0.89 $\pm$ 0.22 & \cellcolor{TMidBg}0.00 $\pm$ 0.00 & \cellcolor{TMidBg}0.63 $\pm$ 0.02 & \cellcolor{TMidBg}0.85 $\pm$ 0.25 & \cellcolor{TMidBg}0.01 $\pm$ 0.01 \\
\cellcolor{TMidBg} & \cellcolor{TMidBg}\multirow{-6}{*}{\begin{tabular}[c]{@{}l@{}}Qwen/Qwen3-Next\\80B-A3B-Instruct\\FP8\end{tabular}} & \cellcolor{TMidBg}\multirow{-2}{*}{\begin{tabular}[c]{@{}c@{}}Reference Guided\\Grading\end{tabular}} & \cellcolor{TMidBg}CoT & \cellcolor{TMidBg}0.57 $\pm$ 0.02 & \cellcolor{TMidBg}0.84 $\pm$ 0.24 & \cellcolor{TMidBg}0.08 $\pm$ 0.06 & \cellcolor{TMidBg}0.60 $\pm$ 0.04 & \cellcolor{TMidBg}0.82 $\pm$ 0.25 & \cellcolor{TMidBg}0.15 $\pm$ 0.07 \\
\cellcolor{TMidBg} & \cellcolor{TMidBg} & \cellcolor{TMidBg} & \cellcolor{TMidBg}Direct & \cellcolor{TMidBg}0.59 $\pm$ 0.00 & \cellcolor{TMidBg}\textbf{0.99 $\pm$ 0.09} & \cellcolor{TMidBg}0.00 $\pm$ 0.00 & \cellcolor{TMidBg}0.60 $\pm$ 0.00 & \cellcolor{TMidBg}\textbf{0.98 $\pm$ 0.12} & \cellcolor{TMidBg}0.00 $\pm$ 0.00 \\
\cellcolor{TMidBg} & \cellcolor{TMidBg} & \cellcolor{TMidBg}\multirow{-2}{*}{\begin{tabular}[c]{@{}c@{}}Pairwise\\Comparison\end{tabular}} & \cellcolor{TMidBg}CoT & \cellcolor{TMidBg}0.56 $\pm$ 0.01 & \cellcolor{TMidBg}0.96 $\pm$ 0.12 & \cellcolor{TMidBg}0.00 $\pm$ 0.00 & \cellcolor{TMidBg}0.54 $\pm$ 0.01 & \cellcolor{TMidBg}0.96 $\pm$ 0.14 & \cellcolor{TMidBg}0.00 $\pm$ 0.00 \\
\cellcolor{TMidBg} & \cellcolor{TMidBg} & \cellcolor{TMidBg} & \cellcolor{TMidBg}Direct & \cellcolor{TMidBg}0.50 $\pm$ 0.01 & \cellcolor{TMidBg}0.87 $\pm$ 0.22 & \cellcolor{TMidBg}0.00 $\pm$ 0.00 & \cellcolor{TMidBg}0.56 $\pm$ 0.02 & \cellcolor{TMidBg}0.83 $\pm$ 0.24 & \cellcolor{TMidBg}0.00 $\pm$ 0.00 \\
\cellcolor{TMidBg} & \cellcolor{TMidBg} & \cellcolor{TMidBg}\multirow{-2}{*}{\begin{tabular}[c]{@{}c@{}}Single Answer\\Grading\end{tabular}} & \cellcolor{TMidBg}CoT & \cellcolor{TMidBg}0.48 $\pm$ 0.02 & \cellcolor{TMidBg}0.82 $\pm$ 0.22 & \cellcolor{TMidBg}0.00 $\pm$ 0.00 & \cellcolor{TMidBg}0.61 $\pm$ 0.03 & \cellcolor{TMidBg}0.78 $\pm$ 0.26 & \cellcolor{TMidBg}0.00 $\pm$ 0.00 \\
\cellcolor{TMidBg} & \cellcolor{TMidBg} & \cellcolor{TMidBg} & \cellcolor{TMidBg}Direct & \cellcolor{TMidBg}0.57 $\pm$ 0.01 & \cellcolor{TMidBg}0.98 $\pm$ 0.11 & \cellcolor{TMidBg}0.00 $\pm$ 0.00 & \cellcolor{TMidBg}0.68 $\pm$ 0.01 & \cellcolor{TMidBg}0.96 $\pm$ 0.15 & \cellcolor{TMidBg}0.00 $\pm$ 0.00 \\
\cellcolor{TMidBg} & \cellcolor{TMidBg}\multirow{-6}{*}{\begin{tabular}[c]{@{}l@{}}google/\\gemma-3-27b-it\end{tabular}} & \cellcolor{TMidBg}\multirow{-2}{*}{\begin{tabular}[c]{@{}c@{}}Reference Guided\\Grading\end{tabular}} & \cellcolor{TMidBg}CoT & \cellcolor{TMidBg}0.57 $\pm$ 0.01 & \cellcolor{TMidBg}0.90 $\pm$ 0.20 & \cellcolor{TMidBg}0.00 $\pm$ 0.00 & \cellcolor{TMidBg}0.68 $\pm$ 0.02 & \cellcolor{TMidBg}0.85 $\pm$ 0.24 & \cellcolor{TMidBg}0.00 $\pm$ 0.00 \\
\cellcolor{TMidBg} & \cellcolor{TMidBg} & \cellcolor{TMidBg} & \cellcolor{TMidBg}Direct & \cellcolor{TMidBg}0.63 $\pm$ 0.08 & \cellcolor{TMidBg}0.83 $\pm$ 0.32 & \cellcolor{TMidBg}0.81 $\pm$ 0.02 & \cellcolor{TMidBg}0.69 $\pm$ 0.08 & \cellcolor{TMidBg}0.77 $\pm$ 0.36 & \cellcolor{TMidBg}0.79 $\pm$ 0.03 \\
\cellcolor{TMidBg} & \cellcolor{TMidBg} & \cellcolor{TMidBg}\multirow{-2}{*}{\begin{tabular}[c]{@{}c@{}}Pairwise\\Comparison\end{tabular}} & \cellcolor{TMidBg}CoT & \cellcolor{TMidBg}0.64 $\pm$ 0.03 & \cellcolor{TMidBg}0.86 $\pm$ 0.29 & \cellcolor{TMidBg}0.69 $\pm$ 0.02 & \cellcolor{TMidBg}0.66 $\pm$ 0.06 & \cellcolor{TMidBg}0.79 $\pm$ 0.33 & \cellcolor{TMidBg}0.72 $\pm$ 0.03 \\
\cellcolor{TMidBg} & \cellcolor{TMidBg} & \cellcolor{TMidBg} & \cellcolor{TMidBg}Direct & \cellcolor{TMidBg}0.58 $\pm$ 0.20 & \cellcolor{TMidBg}0.48 $\pm$ 0.45 & \cellcolor{TMidBg}0.98 $\pm$ 0.01 & \cellcolor{TMidBg}0.65 $\pm$ 0.34 & \cellcolor{TMidBg}0.47 $\pm$ 0.51 & \cellcolor{TMidBg}0.98 $\pm$ 0.01 \\
\cellcolor{TMidBg} & \cellcolor{TMidBg} & \cellcolor{TMidBg}\multirow{-2}{*}{\begin{tabular}[c]{@{}c@{}}Single Answer\\Grading\end{tabular}} & \cellcolor{TMidBg}CoT & \cellcolor{TMidBg}0.58 $\pm$ 0.06 & \cellcolor{TMidBg}0.73 $\pm$ 0.38 & \cellcolor{TMidBg}0.84 $\pm$ 0.02 & \cellcolor{TMidBg}0.68 $\pm$ 0.08 & \cellcolor{TMidBg}0.66 $\pm$ 0.43 & \cellcolor{TMidBg}0.86 $\pm$ 0.02 \\
\cellcolor{TMidBg} & \cellcolor{TMidBg} & \cellcolor{TMidBg} & \cellcolor{TMidBg}Direct & \cellcolor{TMidBg}\textbf{0.69 $\pm$ 0.08} & \cellcolor{TMidBg}0.89 $\pm$ 0.28 & \cellcolor{TMidBg}0.97 $\pm$ 0.01 & \cellcolor{TMidBg}0.63 $\pm$ 0.27 & \cellcolor{TMidBg}0.74 $\pm$ 0.39 & \cellcolor{TMidBg}0.97 $\pm$ 0.01 \\
\cellcolor{TMidBg}\multirow{-18}{*}{\textbf{\textcolor{TMidText}{\faThermometerHalf} 1 \textcolor{TMidText}{\faThermometerHalf}}} & \cellcolor{TMidBg}\multirow{-6}{*}{\begin{tabular}[c]{@{}l@{}}Qwen3.8-Flash-Next\\FP8\end{tabular}} & \cellcolor{TMidBg}\multirow{-2}{*}{\begin{tabular}[c]{@{}c@{}}Reference Guided\\Grading\end{tabular}} & \cellcolor{TMidBg}CoT & \cellcolor{TMidBg}0.62 $\pm$ 0.07 & \cellcolor{TMidBg}0.83 $\pm$ 0.33 & \cellcolor{TMidBg}0.89 $\pm$ 0.01 & \cellcolor{TMidBg}\textbf{0.76 $\pm$ 0.10} & \cellcolor{TMidBg}0.78 $\pm$ 0.37 & \cellcolor{TMidBg}0.89 $\pm$ 0.03 \\

\midrule

\cellcolor{TMildBg} & \cellcolor{TMildBg} & \cellcolor{TMildBg} & \cellcolor{TMildBg}Direct & \cellcolor{TMildBg}0.61 $\pm$ 0.01 & \cellcolor{TMildBg}0.94 $\pm$ 0.16 & \cellcolor{TMildBg}0.00 $\pm$ 0.00 & \cellcolor{TMildBg}0.59 $\pm$ 0.02 & \cellcolor{TMildBg}0.89 $\pm$ 0.21 & \cellcolor{TMildBg}0.01 $\pm$ 0.01 \\
\cellcolor{TMildBg} & \cellcolor{TMildBg} & \cellcolor{TMildBg}\multirow{-2}{*}{\begin{tabular}[c]{@{}c@{}}Pairwise\\Comparison\end{tabular}} & \cellcolor{TMildBg}CoT & \cellcolor{TMildBg}0.59 $\pm$ 0.02 & \cellcolor{TMildBg}0.89 $\pm$ 0.20 & \cellcolor{TMildBg}0.08 $\pm$ 0.09 & \cellcolor{TMildBg}0.53 $\pm$ 0.06 & \cellcolor{TMildBg}0.83 $\pm$ 0.24 & \cellcolor{TMildBg}0.29 $\pm$ 0.18 \\
\cellcolor{TMildBg} & \cellcolor{TMildBg} & \cellcolor{TMildBg} & \cellcolor{TMildBg}Direct & \cellcolor{TMildBg}0.53 $\pm$ 0.02 & \cellcolor{TMildBg}0.84 $\pm$ 0.22 & \cellcolor{TMildBg}0.00 $\pm$ 0.00 & \cellcolor{TMildBg}0.63 $\pm$ 0.03 & \cellcolor{TMildBg}0.80 $\pm$ 0.25 & \cellcolor{TMildBg}0.00 $\pm$ 0.00 \\
\cellcolor{TMildBg} & \cellcolor{TMildBg} & \cellcolor{TMildBg}\multirow{-2}{*}{\begin{tabular}[c]{@{}c@{}}Single Answer\\Grading\end{tabular}} & \cellcolor{TMildBg}CoT & \cellcolor{TMildBg}0.53 $\pm$ 0.02 & \cellcolor{TMildBg}0.78 $\pm$ 0.27 & \cellcolor{TMildBg}0.03 $\pm$ 0.01 & \cellcolor{TMildBg}0.58 $\pm$ 0.05 & \cellcolor{TMildBg}0.69 $\pm$ 0.30 & \cellcolor{TMildBg}0.20 $\pm$ 0.10 \\
\cellcolor{TMildBg} & \cellcolor{TMildBg} & \cellcolor{TMildBg} & \cellcolor{TMildBg}Direct & \cellcolor{TMildBg}0.56 $\pm$ 0.02 & \cellcolor{TMildBg}0.85 $\pm$ 0.24 & \cellcolor{TMildBg}0.00 $\pm$ 0.01 & \cellcolor{TMildBg}0.63 $\pm$ 0.02 & \cellcolor{TMildBg}0.80 $\pm$ 0.28 & \cellcolor{TMildBg}0.02 $\pm$ 0.02 \\
\cellcolor{TMildBg} & \cellcolor{TMildBg}\multirow{-6}{*}{\begin{tabular}[c]{@{}l@{}}Qwen/Qwen3-Next\\80B-A3B-Instruct\\FP8\end{tabular}} & \cellcolor{TMildBg}\multirow{-2}{*}{\begin{tabular}[c]{@{}c@{}}Reference Guided\\Grading\end{tabular}} & \cellcolor{TMildBg}CoT & \cellcolor{TMildBg}0.56 $\pm$ 0.03 & \cellcolor{TMildBg}0.82 $\pm$ 0.24 & \cellcolor{TMildBg}0.11 $\pm$ 0.09 & \cellcolor{TMildBg}0.59 $\pm$ 0.07 & \cellcolor{TMildBg}0.77 $\pm$ 0.27 & \cellcolor{TMildBg}0.20 $\pm$ 0.14 \\
\cellcolor{TMildBg} & \cellcolor{TMildBg} & \cellcolor{TMildBg} & \cellcolor{TMildBg}Direct & \cellcolor{TMildBg}0.59 $\pm$ 0.00 & \cellcolor{TMildBg}\textbf{0.97 $\pm$ 0.12} & \cellcolor{TMildBg}0.00 $\pm$ 0.00 & \cellcolor{TMildBg}0.61 $\pm$ 0.01 & \cellcolor{TMildBg}\textbf{0.97 $\pm$ 0.13} & \cellcolor{TMildBg}0.00 $\pm$ 0.00 \\
\cellcolor{TMildBg} & \cellcolor{TMildBg} & \cellcolor{TMildBg}\multirow{-2}{*}{\begin{tabular}[c]{@{}c@{}}Pairwise\\Comparison\end{tabular}} & \cellcolor{TMildBg}CoT & \cellcolor{TMildBg}0.56 $\pm$ 0.01 & \cellcolor{TMildBg}0.95 $\pm$ 0.14 & \cellcolor{TMildBg}0.00 $\pm$ 0.00 & \cellcolor{TMildBg}0.56 $\pm$ 0.03 & \cellcolor{TMildBg}0.92 $\pm$ 0.18 & \cellcolor{TMildBg}0.00 $\pm$ 0.00 \\
\cellcolor{TMildBg} & \cellcolor{TMildBg} & \cellcolor{TMildBg} & \cellcolor{TMildBg}Direct & \cellcolor{TMildBg}0.50 $\pm$ 0.01 & \cellcolor{TMildBg}0.84 $\pm$ 0.24 & \cellcolor{TMildBg}0.00 $\pm$ 0.00 & \cellcolor{TMildBg}0.56 $\pm$ 0.02 & \cellcolor{TMildBg}0.75 $\pm$ 0.27 & \cellcolor{TMildBg}0.00 $\pm$ 0.00 \\
\cellcolor{TMildBg} & \cellcolor{TMildBg} & \cellcolor{TMildBg}\multirow{-2}{*}{\begin{tabular}[c]{@{}c@{}}Single Answer\\Grading\end{tabular}} & \cellcolor{TMildBg}CoT & \cellcolor{TMildBg}0.47 $\pm$ 0.01 & \cellcolor{TMildBg}0.78 $\pm$ 0.27 & \cellcolor{TMildBg}0.00 $\pm$ 0.00 & \cellcolor{TMildBg}0.60 $\pm$ 0.02 & \cellcolor{TMildBg}0.74 $\pm$ 0.26 & \cellcolor{TMildBg}0.00 $\pm$ 0.00 \\
\cellcolor{TMildBg} & \cellcolor{TMildBg} & \cellcolor{TMildBg} & \cellcolor{TMildBg}Direct & \cellcolor{TMildBg}0.57 $\pm$ 0.01 & \cellcolor{TMildBg}0.96 $\pm$ 0.13 & \cellcolor{TMildBg}0.00 $\pm$ 0.00 & \cellcolor{TMildBg}0.68 $\pm$ 0.01 & \cellcolor{TMildBg}0.94 $\pm$ 0.19 & \cellcolor{TMildBg}0.00 $\pm$ 0.00 \\
\cellcolor{TMildBg} & \cellcolor{TMildBg}\multirow{-6}{*}{\begin{tabular}[c]{@{}l@{}}google/\\gemma-3-27b-it\end{tabular}} & \cellcolor{TMildBg}\multirow{-2}{*}{\begin{tabular}[c]{@{}c@{}}Reference Guided\\Grading\end{tabular}} & \cellcolor{TMildBg}CoT & \cellcolor{TMildBg}0.57 $\pm$ 0.01 & \cellcolor{TMildBg}0.86 $\pm$ 0.23 & \cellcolor{TMildBg}0.00 $\pm$ 0.00 & \cellcolor{TMildBg}0.67 $\pm$ 0.03 & \cellcolor{TMildBg}0.82 $\pm$ 0.25 & \cellcolor{TMildBg}0.00 $\pm$ 0.00 \\
\cellcolor{TMildBg} & \cellcolor{TMildBg} & \cellcolor{TMildBg} & \cellcolor{TMildBg}Direct & \cellcolor{TMildBg}0.61 $\pm$ 0.06 & \cellcolor{TMildBg}0.81 $\pm$ 0.34 & \cellcolor{TMildBg}0.75 $\pm$ 0.03 & \cellcolor{TMildBg}0.67 $\pm$ 0.08 & \cellcolor{TMildBg}0.77 $\pm$ 0.33 & \cellcolor{TMildBg}0.75 $\pm$ 0.05 \\
\cellcolor{TMildBg} & \cellcolor{TMildBg} & \cellcolor{TMildBg}\multirow{-2}{*}{\begin{tabular}[c]{@{}c@{}}Pairwise\\Comparison\end{tabular}} & \cellcolor{TMildBg}CoT & \cellcolor{TMildBg}0.60 $\pm$ 0.04 & \cellcolor{TMildBg}0.75 $\pm$ 0.35 & \cellcolor{TMildBg}0.64 $\pm$ 0.02 & \cellcolor{TMildBg}0.61 $\pm$ 0.03 & \cellcolor{TMildBg}0.75 $\pm$ 0.34 & \cellcolor{TMildBg}0.67 $\pm$ 0.04 \\
\cellcolor{TMildBg} & \cellcolor{TMildBg} & \cellcolor{TMildBg} & \cellcolor{TMildBg}Direct & \cellcolor{TMildBg}0.52 $\pm$ 0.10 & \cellcolor{TMildBg}0.48 $\pm$ 0.49 & \cellcolor{TMildBg}0.95 $\pm$ 0.02 & \cellcolor{TMildBg}0.53 $\pm$ 0.14 & \cellcolor{TMildBg}0.92 $\pm$ 0.20 & \cellcolor{TMildBg}0.97 $\pm$ 0.01 \\
\cellcolor{TMildBg} & \cellcolor{TMildBg} & \cellcolor{TMildBg}\multirow{-2}{*}{\begin{tabular}[c]{@{}c@{}}Single Answer\\Grading\end{tabular}} & \cellcolor{TMildBg}CoT & \cellcolor{TMildBg}0.58 $\pm$ 0.07 & \cellcolor{TMildBg}0.64 $\pm$ 0.41 & \cellcolor{TMildBg}0.83 $\pm$ 0.02 & \cellcolor{TMildBg}0.59 $\pm$ 0.13 & \cellcolor{TMildBg}0.51 $\pm$ 0.43 & \cellcolor{TMildBg}0.87 $\pm$ 0.03 \\
\cellcolor{TMildBg} & \cellcolor{TMildBg} & \cellcolor{TMildBg} & \cellcolor{TMildBg}Direct & \cellcolor{TMildBg}\textbf{0.62 $\pm$ 0.08} & \cellcolor{TMildBg}0.79 $\pm$ 0.37 & \cellcolor{TMildBg}0.93 $\pm$ 0.01 & \cellcolor{TMildBg}0.56 $\pm$ 0.22 & \cellcolor{TMildBg}0.62 $\pm$ 0.44 & \cellcolor{TMildBg}0.95 $\pm$ 0.03 \\
\cellcolor{TMildBg}\multirow{-18}{*}{\textbf{\textcolor{TMildText}{\faThermometerThreeQuarters} 1.5 \textcolor{TMildText}{\faThermometerThreeQuarters}}} & \cellcolor{TMildBg}\multirow{-6}{*}{\begin{tabular}[c]{@{}l@{}}Qwen3.8-Flash-Next\\FP8\end{tabular}} & \cellcolor{TMildBg}\multirow{-2}{*}{\begin{tabular}[c]{@{}c@{}}Reference Guided\\Grading\end{tabular}} & \cellcolor{TMildBg}CoT & \cellcolor{TMildBg}0.58 $\pm$ 0.06 & \cellcolor{TMildBg}0.73 $\pm$ 0.37 & \cellcolor{TMildBg}0.84 $\pm$ 0.01 & \cellcolor{TMildBg}\textbf{0.69 $\pm$ 0.09} & \cellcolor{TMildBg}0.74 $\pm$ 0.38 & \cellcolor{TMildBg}0.85 $\pm$ 0.03 \\
\bottomrule
\end{tabular}
\end{adjustbox}
\caption{Full results of LLM-as-a-Judge stability for $T \in \{1, 1.5\}$ (part 2 of 3; all temperatures $T \in \{0.01, 0.5, 1.0, 1.5, 2.0, 3.0\}$ are reported across Tables~\ref{tab:appendix_full_results_1}--\ref{tab:appendix_full_results_3}). Three judge models, three evaluation paradigms (Pairwise Comparison, Single Answer Grading, and Reference-Guided Grading), and two prompting strategies (Direct and CoT), measured by agreement with human judgments (AGR; $\mu \pm$ std over 10 repeats), 1-flip consistency across repeats (CON; $\mu \pm$ std over questions), and format error rate (ERR; $\mu \pm$ std over 10 repeats) on MT-Bench and MMLU-Pro. \textbf{Bold} marks the best AGR/CON (highest) and the uniquely lowest ERR within each temperature block and dataset.}
\label{tab:appendix_full_results_2}
\end{table*}

\begin{table*}[htbp]
\centering
\begin{adjustbox}{width=1.00\textwidth}
\begin{tabular}{c>{\raggedright\arraybackslash}p{3.25cm}cccccccc}
\toprule
\multirow{2}{*}{\textbf{\faBalanceScale T \faBalanceScale}} & \multirow{2}{*}{\textbf{Model}} & \multirow{2}{*}{\textbf{Judge Type}} & \multirow{2}{*}{\textbf{\begin{tabular}[c]{@{}c@{}}Prompt\\ Strategy\end{tabular}}} & \multicolumn{3}{c}{\textbf{MT-Bench Human Judgments}} & \multicolumn{3}{c}{\textbf{MMLU\_Pro}} \\
\cmidrule(lr){5-7}\cmidrule(lr){8-10}
& & & & \textbf{\begin{tabular}[c]{@{}c@{}}AGR$\uparrow$ \\ ($\mu \pm$ std)\end{tabular}} & \textbf{\begin{tabular}[c]{@{}c@{}}CON$\uparrow$ \\ ($\mu \pm$ std)\end{tabular}} & \textbf{\begin{tabular}[c]{@{}c@{}}ERR$\downarrow$ \\ ($\mu \pm$ std)\end{tabular}} & \textbf{\begin{tabular}[c]{@{}c@{}}AGR$\uparrow$ \\ ($\mu \pm$ std)\end{tabular}} & \textbf{\begin{tabular}[c]{@{}c@{}}CON$\uparrow$ \\ ($\mu \pm$ std)\end{tabular}} & \textbf{\begin{tabular}[c]{@{}c@{}}ERR$\downarrow$ \\ ($\mu \pm$ std)\end{tabular}} \\
\midrule

\cellcolor{TWarmBg} & \cellcolor{TWarmBg} & \cellcolor{TWarmBg} & \cellcolor{TWarmBg}Direct & \cellcolor{TWarmBg}\textbf{0.61 $\pm$ 0.01} & \cellcolor{TWarmBg}0.91 $\pm$ 0.19 & \cellcolor{TWarmBg}0.00 $\pm$ 0.01 & \cellcolor{TWarmBg}0.58 $\pm$ 0.03 & \cellcolor{TWarmBg}0.84 $\pm$ 0.24 & \cellcolor{TWarmBg}0.03 $\pm$ 0.05 \\
\cellcolor{TWarmBg} & \cellcolor{TWarmBg} & \cellcolor{TWarmBg}\multirow{-2}{*}{\begin{tabular}[c]{@{}c@{}}Pairwise\\Comparison\end{tabular}} & \cellcolor{TWarmBg}CoT & \cellcolor{TWarmBg}0.57 $\pm$ 0.03 & \cellcolor{TWarmBg}0.85 $\pm$ 0.22 & \cellcolor{TWarmBg}0.14 $\pm$ 0.13 & \cellcolor{TWarmBg}0.51 $\pm$ 0.08 & \cellcolor{TWarmBg}0.80 $\pm$ 0.26 & \cellcolor{TWarmBg}0.35 $\pm$ 0.22 \\
\cellcolor{TWarmBg} & \cellcolor{TWarmBg} & \cellcolor{TWarmBg} & \cellcolor{TWarmBg}Direct & \cellcolor{TWarmBg}0.53 $\pm$ 0.02 & \cellcolor{TWarmBg}0.82 $\pm$ 0.23 & \cellcolor{TWarmBg}0.00 $\pm$ 0.00 & \cellcolor{TWarmBg}0.61 $\pm$ 0.03 & \cellcolor{TWarmBg}0.78 $\pm$ 0.26 & \cellcolor{TWarmBg}0.00 $\pm$ 0.00 \\
\cellcolor{TWarmBg} & \cellcolor{TWarmBg} & \cellcolor{TWarmBg}\multirow{-2}{*}{\begin{tabular}[c]{@{}c@{}}Single Answer\\Grading\end{tabular}} & \cellcolor{TWarmBg}CoT & \cellcolor{TWarmBg}0.52 $\pm$ 0.01 & \cellcolor{TWarmBg}0.75 $\pm$ 0.27 & \cellcolor{TWarmBg}0.04 $\pm$ 0.03 & \cellcolor{TWarmBg}0.59 $\pm$ 0.03 & \cellcolor{TWarmBg}0.69 $\pm$ 0.28 & \cellcolor{TWarmBg}0.22 $\pm$ 0.11 \\
\cellcolor{TWarmBg} & \cellcolor{TWarmBg} & \cellcolor{TWarmBg} & \cellcolor{TWarmBg}Direct & \cellcolor{TWarmBg}0.55 $\pm$ 0.03 & \cellcolor{TWarmBg}0.81 $\pm$ 0.25 & \cellcolor{TWarmBg}0.02 $\pm$ 0.03 & \cellcolor{TWarmBg}0.63 $\pm$ 0.03 & \cellcolor{TWarmBg}0.76 $\pm$ 0.29 & \cellcolor{TWarmBg}0.04 $\pm$ 0.05 \\
\cellcolor{TWarmBg} & \cellcolor{TWarmBg}\multirow{-6}{*}{\begin{tabular}[c]{@{}l@{}}Qwen/Qwen3-Next\\80B-A3B-Instruct\\FP8\end{tabular}} & \cellcolor{TWarmBg}\multirow{-2}{*}{\begin{tabular}[c]{@{}c@{}}Reference Guided\\Grading\end{tabular}} & \cellcolor{TWarmBg}CoT & \cellcolor{TWarmBg}0.55 $\pm$ 0.05 & \cellcolor{TWarmBg}0.76 $\pm$ 0.26 & \cellcolor{TWarmBg}0.18 $\pm$ 0.15 & \cellcolor{TWarmBg}0.54 $\pm$ 0.10 & \cellcolor{TWarmBg}0.75 $\pm$ 0.26 & \cellcolor{TWarmBg}0.27 $\pm$ 0.19 \\
\cellcolor{TWarmBg} & \cellcolor{TWarmBg} & \cellcolor{TWarmBg} & \cellcolor{TWarmBg}Direct & \cellcolor{TWarmBg}0.58 $\pm$ 0.00 & \cellcolor{TWarmBg}\textbf{0.96 $\pm$ 0.14} & \cellcolor{TWarmBg}0.00 $\pm$ 0.00 & \cellcolor{TWarmBg}0.60 $\pm$ 0.01 & \cellcolor{TWarmBg}\textbf{0.96 $\pm$ 0.15} & \cellcolor{TWarmBg}0.00 $\pm$ 0.00 \\
\cellcolor{TWarmBg} & \cellcolor{TWarmBg} & \cellcolor{TWarmBg}\multirow{-2}{*}{\begin{tabular}[c]{@{}c@{}}Pairwise\\Comparison\end{tabular}} & \cellcolor{TWarmBg}CoT & \cellcolor{TWarmBg}0.56 $\pm$ 0.01 & \cellcolor{TWarmBg}0.92 $\pm$ 0.17 & \cellcolor{TWarmBg}0.00 $\pm$ 0.01 & \cellcolor{TWarmBg}0.55 $\pm$ 0.02 & \cellcolor{TWarmBg}0.87 $\pm$ 0.23 & \cellcolor{TWarmBg}0.01 $\pm$ 0.01 \\
\cellcolor{TWarmBg} & \cellcolor{TWarmBg} & \cellcolor{TWarmBg} & \cellcolor{TWarmBg}Direct & \cellcolor{TWarmBg}0.50 $\pm$ 0.02 & \cellcolor{TWarmBg}0.80 $\pm$ 0.25 & \cellcolor{TWarmBg}0.00 $\pm$ 0.00 & \cellcolor{TWarmBg}0.55 $\pm$ 0.04 & \cellcolor{TWarmBg}0.70 $\pm$ 0.27 & \cellcolor{TWarmBg}0.00 $\pm$ 0.00 \\
\cellcolor{TWarmBg} & \cellcolor{TWarmBg} & \cellcolor{TWarmBg}\multirow{-2}{*}{\begin{tabular}[c]{@{}c@{}}Single Answer\\Grading\end{tabular}} & \cellcolor{TWarmBg}CoT & \cellcolor{TWarmBg}0.48 $\pm$ 0.02 & \cellcolor{TWarmBg}0.71 $\pm$ 0.28 & \cellcolor{TWarmBg}0.00 $\pm$ 0.00 & \cellcolor{TWarmBg}0.58 $\pm$ 0.04 & \cellcolor{TWarmBg}0.69 $\pm$ 0.26 & \cellcolor{TWarmBg}0.01 $\pm$ 0.01 \\
\cellcolor{TWarmBg} & \cellcolor{TWarmBg} & \cellcolor{TWarmBg} & \cellcolor{TWarmBg}Direct & \cellcolor{TWarmBg}0.57 $\pm$ 0.01 & \cellcolor{TWarmBg}0.93 $\pm$ 0.17 & \cellcolor{TWarmBg}0.00 $\pm$ 0.00 & \cellcolor{TWarmBg}\textbf{0.68 $\pm$ 0.01} & \cellcolor{TWarmBg}0.92 $\pm$ 0.21 & \cellcolor{TWarmBg}0.00 $\pm$ 0.00 \\
\cellcolor{TWarmBg} & \cellcolor{TWarmBg}\multirow{-6}{*}{\begin{tabular}[c]{@{}l@{}}google/\\gemma-3-27b-it\end{tabular}} & \cellcolor{TWarmBg}\multirow{-2}{*}{\begin{tabular}[c]{@{}c@{}}Reference Guided\\Grading\end{tabular}} & \cellcolor{TWarmBg}CoT & \cellcolor{TWarmBg}0.56 $\pm$ 0.02 & \cellcolor{TWarmBg}0.84 $\pm$ 0.23 & \cellcolor{TWarmBg}0.02 $\pm$ 0.01 & \cellcolor{TWarmBg}0.67 $\pm$ 0.02 & \cellcolor{TWarmBg}0.76 $\pm$ 0.27 & \cellcolor{TWarmBg}0.01 $\pm$ 0.01 \\
\cellcolor{TWarmBg} & \cellcolor{TWarmBg} & \cellcolor{TWarmBg} & \cellcolor{TWarmBg}Direct & \cellcolor{TWarmBg}0.56 $\pm$ 0.05 & \cellcolor{TWarmBg}0.71 $\pm$ 0.38 & \cellcolor{TWarmBg}0.74 $\pm$ 0.04 & \cellcolor{TWarmBg}0.58 $\pm$ 0.05 & \cellcolor{TWarmBg}0.62 $\pm$ 0.40 & \cellcolor{TWarmBg}0.73 $\pm$ 0.04 \\
\cellcolor{TWarmBg} & \cellcolor{TWarmBg} & \cellcolor{TWarmBg}\multirow{-2}{*}{\begin{tabular}[c]{@{}c@{}}Pairwise\\Comparison\end{tabular}} & \cellcolor{TWarmBg}CoT & \cellcolor{TWarmBg}0.47 $\pm$ 0.05 & \cellcolor{TWarmBg}0.63 $\pm$ 0.37 & \cellcolor{TWarmBg}0.71 $\pm$ 0.02 & \cellcolor{TWarmBg}0.41 $\pm$ 0.07 & \cellcolor{TWarmBg}0.53 $\pm$ 0.40 & \cellcolor{TWarmBg}0.76 $\pm$ 0.04 \\
\cellcolor{TWarmBg} & \cellcolor{TWarmBg} & \cellcolor{TWarmBg} & \cellcolor{TWarmBg}Direct & \cellcolor{TWarmBg}0.48 $\pm$ 0.12 & \cellcolor{TWarmBg}0.37 $\pm$ 0.43 & \cellcolor{TWarmBg}0.93 $\pm$ 0.02 & \cellcolor{TWarmBg}0.56 $\pm$ 0.24 & \cellcolor{TWarmBg}0.40 $\pm$ 0.52 & \cellcolor{TWarmBg}0.96 $\pm$ 0.02 \\
\cellcolor{TWarmBg} & \cellcolor{TWarmBg} & \cellcolor{TWarmBg}\multirow{-2}{*}{\begin{tabular}[c]{@{}c@{}}Single Answer\\Grading\end{tabular}} & \cellcolor{TWarmBg}CoT & \cellcolor{TWarmBg}0.38 $\pm$ 0.10 & \cellcolor{TWarmBg}0.41 $\pm$ 0.42 & \cellcolor{TWarmBg}0.86 $\pm$ 0.03 & \cellcolor{TWarmBg}0.35 $\pm$ 0.15 & \cellcolor{TWarmBg}0.32 $\pm$ 0.41 & \cellcolor{TWarmBg}0.92 $\pm$ 0.02 \\
\cellcolor{TWarmBg} & \cellcolor{TWarmBg} & \cellcolor{TWarmBg} & \cellcolor{TWarmBg}Direct & \cellcolor{TWarmBg}0.56 $\pm$ 0.07 & \cellcolor{TWarmBg}0.67 $\pm$ 0.41 & \cellcolor{TWarmBg}0.89 $\pm$ 0.02 & \cellcolor{TWarmBg}0.63 $\pm$ 0.13 & \cellcolor{TWarmBg}0.57 $\pm$ 0.47 & \cellcolor{TWarmBg}0.91 $\pm$ 0.03 \\
\cellcolor{TWarmBg}\multirow{-18}{*}{\textbf{\textcolor{TWarmText}{\faThermometerFull} 2 \textcolor{TWarmText}{\faThermometerFull}}} & \cellcolor{TWarmBg}\multirow{-6}{*}{\begin{tabular}[c]{@{}l@{}}Qwen3.8-Flash-Next\\FP8\end{tabular}} & \cellcolor{TWarmBg}\multirow{-2}{*}{\begin{tabular}[c]{@{}c@{}}Reference Guided\\Grading\end{tabular}} & \cellcolor{TWarmBg}CoT & \cellcolor{TWarmBg}0.48 $\pm$ 0.08 & \cellcolor{TWarmBg}0.64 $\pm$ 0.41 & \cellcolor{TWarmBg}0.85 $\pm$ 0.02 & \cellcolor{TWarmBg}0.51 $\pm$ 0.15 & \cellcolor{TWarmBg}0.55 $\pm$ 0.45 & \cellcolor{TWarmBg}0.88 $\pm$ 0.02 \\

\midrule

\cellcolor{THotBg} & \cellcolor{THotBg} & \cellcolor{THotBg} & \cellcolor{THotBg}Direct & \cellcolor{THotBg}\textbf{0.59 $\pm$ 0.03} & \cellcolor{THotBg}0.84 $\pm$ 0.23 & \cellcolor{THotBg}0.07 $\pm$ 0.07 & \cellcolor{THotBg}0.54 $\pm$ 0.06 & \cellcolor{THotBg}0.68 $\pm$ 0.26 & \cellcolor{THotBg}0.17 $\pm$ 0.19 \\
\cellcolor{THotBg} & \cellcolor{THotBg} & \cellcolor{THotBg}\multirow{-2}{*}{\begin{tabular}[c]{@{}c@{}}Pairwise\\Comparison\end{tabular}} & \cellcolor{THotBg}CoT & \cellcolor{THotBg}0.53 $\pm$ 0.07 & \cellcolor{THotBg}0.66 $\pm$ 0.33 & \cellcolor{THotBg}0.37 $\pm$ 0.13 & \cellcolor{THotBg}0.45 $\pm$ 0.10 & \cellcolor{THotBg}0.61 $\pm$ 0.33 & \cellcolor{THotBg}0.46 $\pm$ 0.15 \\
\cellcolor{THotBg} & \cellcolor{THotBg} & \cellcolor{THotBg} & \cellcolor{THotBg}Direct & \cellcolor{THotBg}0.52 $\pm$ 0.03 & \cellcolor{THotBg}0.76 $\pm$ 0.25 & \cellcolor{THotBg}0.00 $\pm$ 0.00 & \cellcolor{THotBg}0.60 $\pm$ 0.02 & \cellcolor{THotBg}0.72 $\pm$ 0.25 & \cellcolor{THotBg}0.00 $\pm$ 0.00 \\
\cellcolor{THotBg} & \cellcolor{THotBg} & \cellcolor{THotBg}\multirow{-2}{*}{\begin{tabular}[c]{@{}c@{}}Single Answer\\Grading\end{tabular}} & \cellcolor{THotBg}CoT & \cellcolor{THotBg}0.50 $\pm$ 0.02 & \cellcolor{THotBg}0.63 $\pm$ 0.29 & \cellcolor{THotBg}0.13 $\pm$ 0.04 & \cellcolor{THotBg}0.51 $\pm$ 0.04 & \cellcolor{THotBg}0.58 $\pm$ 0.32 & \cellcolor{THotBg}0.35 $\pm$ 0.07 \\
\cellcolor{THotBg} & \cellcolor{THotBg} & \cellcolor{THotBg} & \cellcolor{THotBg}Direct & \cellcolor{THotBg}0.54 $\pm$ 0.06 & \cellcolor{THotBg}0.70 $\pm$ 0.26 & \cellcolor{THotBg}0.13 $\pm$ 0.12 & \cellcolor{THotBg}0.60 $\pm$ 0.06 & \cellcolor{THotBg}0.71 $\pm$ 0.26 & \cellcolor{THotBg}0.17 $\pm$ 0.17 \\
\cellcolor{THotBg} & \cellcolor{THotBg}\multirow{-6}{*}{\begin{tabular}[c]{@{}l@{}}Qwen/Qwen3-Next\\80B-A3B-Instruct\\FP8\end{tabular}} & \cellcolor{THotBg}\multirow{-2}{*}{\begin{tabular}[c]{@{}c@{}}Reference Guided\\Grading\end{tabular}} & \cellcolor{THotBg}CoT & \cellcolor{THotBg}0.45 $\pm$ 0.06 & \cellcolor{THotBg}0.55 $\pm$ 0.30 & \cellcolor{THotBg}0.37 $\pm$ 0.09 & \cellcolor{THotBg}0.44 $\pm$ 0.10 & \cellcolor{THotBg}0.57 $\pm$ 0.32 & \cellcolor{THotBg}0.49 $\pm$ 0.08 \\
\cellcolor{THotBg} & \cellcolor{THotBg} & \cellcolor{THotBg} & \cellcolor{THotBg}Direct & \cellcolor{THotBg}0.58 $\pm$ 0.00 & \cellcolor{THotBg}\textbf{0.93 $\pm$ 0.19} & \cellcolor{THotBg}0.00 $\pm$ 0.00 & \cellcolor{THotBg}0.59 $\pm$ 0.01 & \cellcolor{THotBg}\textbf{0.90 $\pm$ 0.20} & \cellcolor{THotBg}0.00 $\pm$ 0.00 \\
\cellcolor{THotBg} & \cellcolor{THotBg} & \cellcolor{THotBg}\multirow{-2}{*}{\begin{tabular}[c]{@{}c@{}}Pairwise\\Comparison\end{tabular}} & \cellcolor{THotBg}CoT & \cellcolor{THotBg}0.55 $\pm$ 0.02 & \cellcolor{THotBg}0.86 $\pm$ 0.23 & \cellcolor{THotBg}0.07 $\pm$ 0.03 & \cellcolor{THotBg}0.53 $\pm$ 0.03 & \cellcolor{THotBg}0.75 $\pm$ 0.27 & \cellcolor{THotBg}0.09 $\pm$ 0.04 \\
\cellcolor{THotBg} & \cellcolor{THotBg} & \cellcolor{THotBg} & \cellcolor{THotBg}Direct & \cellcolor{THotBg}0.49 $\pm$ 0.01 & \cellcolor{THotBg}0.74 $\pm$ 0.25 & \cellcolor{THotBg}0.00 $\pm$ 0.00 & \cellcolor{THotBg}0.53 $\pm$ 0.05 & \cellcolor{THotBg}0.66 $\pm$ 0.24 & \cellcolor{THotBg}0.00 $\pm$ 0.00 \\
\cellcolor{THotBg} & \cellcolor{THotBg} & \cellcolor{THotBg}\multirow{-2}{*}{\begin{tabular}[c]{@{}c@{}}Single Answer\\Grading\end{tabular}} & \cellcolor{THotBg}CoT & \cellcolor{THotBg}0.47 $\pm$ 0.02 & \cellcolor{THotBg}0.63 $\pm$ 0.27 & \cellcolor{THotBg}0.01 $\pm$ 0.01 & \cellcolor{THotBg}0.56 $\pm$ 0.03 & \cellcolor{THotBg}0.57 $\pm$ 0.27 & \cellcolor{THotBg}0.03 $\pm$ 0.01 \\
\cellcolor{THotBg} & \cellcolor{THotBg} & \cellcolor{THotBg} & \cellcolor{THotBg}Direct & \cellcolor{THotBg}0.56 $\pm$ 0.01 & \cellcolor{THotBg}0.88 $\pm$ 0.21 & \cellcolor{THotBg}0.00 $\pm$ 0.00 & \cellcolor{THotBg}\textbf{0.67 $\pm$ 0.03} & \cellcolor{THotBg}0.86 $\pm$ 0.25 & \cellcolor{THotBg}0.00 $\pm$ 0.00 \\
\cellcolor{THotBg}\multirow{-12}{*}{\textbf{\textcolor{THotText}{\faFire} 3 \textcolor{THotText}{\faFire}}} & \cellcolor{THotBg}\multirow{-6}{*}{\begin{tabular}[c]{@{}l@{}}google/\\gemma-3-27b-it\end{tabular}} & \cellcolor{THotBg}\multirow{-2}{*}{\begin{tabular}[c]{@{}c@{}}Reference Guided\\Grading\end{tabular}} & \cellcolor{THotBg}CoT & \cellcolor{THotBg}0.55 $\pm$ 0.02 & \cellcolor{THotBg}0.73 $\pm$ 0.27 & \cellcolor{THotBg}0.10 $\pm$ 0.02 & \cellcolor{THotBg}0.65 $\pm$ 0.04 & \cellcolor{THotBg}0.66 $\pm$ 0.29 & \cellcolor{THotBg}0.10 $\pm$ 0.04 \\
\bottomrule
\end{tabular}
\end{adjustbox}
\caption{Full results of LLM-as-a-Judge stability for $T \in \{2, 3\}$ (part 3 of 3; all temperatures $T \in \{0.01, 0.5, 1.0, 1.5, 2.0, 3.0\}$ are reported across Tables~\ref{tab:appendix_full_results_1}--\ref{tab:appendix_full_results_3}). Three judge models, three evaluation paradigms (Pairwise Comparison, Single Answer Grading, and Reference-Guided Grading), and two prompting strategies (Direct and CoT), measured by agreement with human judgments (AGR; $\mu \pm$ std over 10 repeats), 1-flip consistency across repeats (CON; $\mu \pm$ std over questions), and format error rate (ERR; $\mu \pm$ std over 10 repeats) on MT-Bench and MMLU-Pro. \textbf{Bold} marks the best AGR/CON (highest) and the uniquely lowest ERR within each temperature block and dataset. Qwen3.8-Flash-Next-FP8 was not evaluated at $T=3.0$.}
\label{tab:appendix_full_results_3}
\end{table*}

\end{document}